\pdfoutput=1
\documentclass[11pt,a4paper]{article}

\usepackage[T1]{fontenc}
\usepackage[utf8]{inputenc}
\IfFileExists{lmodern.sty}{\usepackage{lmodern}\usepackage{microtype}}%
                          {\usepackage[protrusion=true,expansion=false]{microtype}}
\usepackage[margin=2.4cm,top=2.5cm,bottom=2.5cm]{geometry}
\usepackage{setspace}
\usepackage{amsmath,amssymb}
\usepackage{graphicx}
\graphicspath{{figures/}}
\usepackage{booktabs,longtable,tabularx,array,multirow}
\usepackage{caption}
\usepackage{float}
\newcolumntype{L}[1]{>{\raggedright\arraybackslash}p{#1}}
\usepackage[dvipsnames,table]{xcolor}
\usepackage{enumitem}
\setlist{itemsep=1pt,topsep=3pt,parsep=0pt}
\usepackage{xltabular}      
\usepackage{seqsplit}       
\usepackage{titlesec}
\usepackage[numbers,sort&compress]{natbib}
\usepackage[hidelinks,breaklinks=true]{hyperref}
\hypersetup{pdftitle={The widening evaluation gap in medical large language model research 2023 to 2026},
            pdfauthor={Raad Bin Tareaf, Murad Al-Rajab, Samia Loucif},
            pdfkeywords={large language models, evidence map, meta-research, reproducibility, digital medicine}}

\title{\vspace{-1.4cm}\bfseries The widening evaluation gap in medical large
language model research 2023 to 2026}
\author{}
\date{}

\begin{document}
\maketitle
\vspace{-2.2cm}

\begin{center}
\begin{minipage}{0.94\textwidth}\centering
Raad Bin Tareaf\textsuperscript{1,*}, Murad Al-Rajab\textsuperscript{2},
Samia Loucif\textsuperscript{3}

\smallskip
{\small
\textsuperscript{1}XU Exponential University of Applied Sciences, Data Science
and Artificial Intelligence Cluster, Marlene-Dietrich-Allee 12B, 14482 Potsdam,
Germany\\
\textsuperscript{2}College of Engineering, Abu Dhabi University, Abu Dhabi,
United Arab Emirates\\
\textsuperscript{3}College of Technological Innovation, Zayed University,
Abu Dhabi, United Arab Emirates\\[2pt]
\textsuperscript{*}Correspondence: \texttt{r.bintareaf@xu-university.de}}
\end{minipage}
\end{center}

\vspace{6pt}
\hrule
\vspace{8pt}

\noindent\textbf{Abstract.}
Large language models are superseded every few quarters; clinical evidence
takes years. We asked whether medical research is keeping pace with the systems
it evaluates. PubMed returned 11,628 records for January 2023 to June 2026
across fourteen clinical domains, growing 45-fold; 2.5\% used a randomised,
controlled or prospective design. Evaluation lag, from a study's newest named
model release to its own publication, widened from 1.33 to 6.08 quarters.
Because discontinued models age mechanically, we benchmarked this against a
counterfactual holding model composition fixed: migration to newer systems
offset only 56\% of the drift (95\% CI 50--65). Randomised trials evaluated
models a median 4.6 quarters older than other designs
($P=3\times10^{-19}$), yet among studies naming a model still under development
no design differed from any other; 62\% of randomised trials evaluated a
discontinued family. Rigour and currency are in tension, and that tension
reflects model selection rather than research timelines.

\vspace{6pt}
\hrule
\vspace{10pt}

\section{Introduction}

Large language models entered medicine faster than any preceding class of
clinical software. Where the previous generation of clinical natural language
processing was built on task-specific encoders trained for a single
purpose\cite{Zhang2022}, a single general-purpose model is now applied across
diagnosis, documentation and triage without retraining. Within three years of
the release of ChatGPT, models had matched or exceeded clinician performance
on licensing examinations\cite{Singhal2025,Liu2024,Kanjee2023} and approached
specialist performance on differential diagnosis\cite{McDuff2025}. They were
being evaluated for diagnosis, documentation, triage, education and
pharmacovigilance\cite{Bedi2025,Meng2024}, and the resulting literature now
exceeds ten thousand indexed reports. The pace of that expansion is itself
well documented. What has not been examined is whether the evidence base is
keeping up with the technology it describes.

The question matters because the two move on different clocks. Model families
are superseded every six to nine months. Randomised trials take one to two
years from design to publication, and the first randomised evaluations of
these systems in clinical reasoning appeared only in late
2024\cite{Goh2024,Han2024}. If those rates diverge, a structural problem
follows: the most rigorous evidence available at any moment concerns models
that are no longer the ones being deployed. Clinicians and regulators would
then be asked to make decisions about current systems using trials of
superseded ones, and no amount of methodological improvement within any single
study would resolve it.

Existing syntheses cannot answer this. Narrative reviews of the field
characterise applications and challenges but do not measure the temporal
relationship between evidence and technology\cite{Wang2024,Meng2024}.
Systematic reviews of individual applications establish what particular models
achieve on particular tasks\cite{Bedi2025,Li2023,Tam2024}, but by construction
examine a slice of the literature at a fixed moment. Neither design produces the quantity that matters here: the
distance, in time, between what a study evaluates and what exists when it
appears.

We address this with a reproducible evidence map of the indexed literature on
generative language models in healthcare, covering January 2023 to June 2026.
Harvesting, screening, classification and analysis execute end to end from
deposited code, so the map can be regenerated as the literature grows. Within
it we define and measure the evaluation lag: the interval between the release
of the newest model a study names and the study's own publication. We then ask
whether that lag differs by study design, whether it is widening, and how it
relates to the reproducibility of model reporting---a property the reporting
guidelines for artificial intelligence in medicine already require, CONSORT-AI
listing algorithm version among the items a trial report must
specify\cite{Liu2020,CruzRivera2020,Vasey2022,Gallifant2025}.

Our aims were to characterise the composition and growth of the evidence base;
to quantify the evaluation lag and test its association with study design and
publication date; to measure how precisely studies specify the models they
evaluate; to describe the structure of the model landscape over time; and to
identify where research attention and trial-grade evidence diverge.

\section{Results}

\subsection*{Composition of the evidence base}

\begin{figure}[H]\centering
\includegraphics[width=0.92\textwidth]{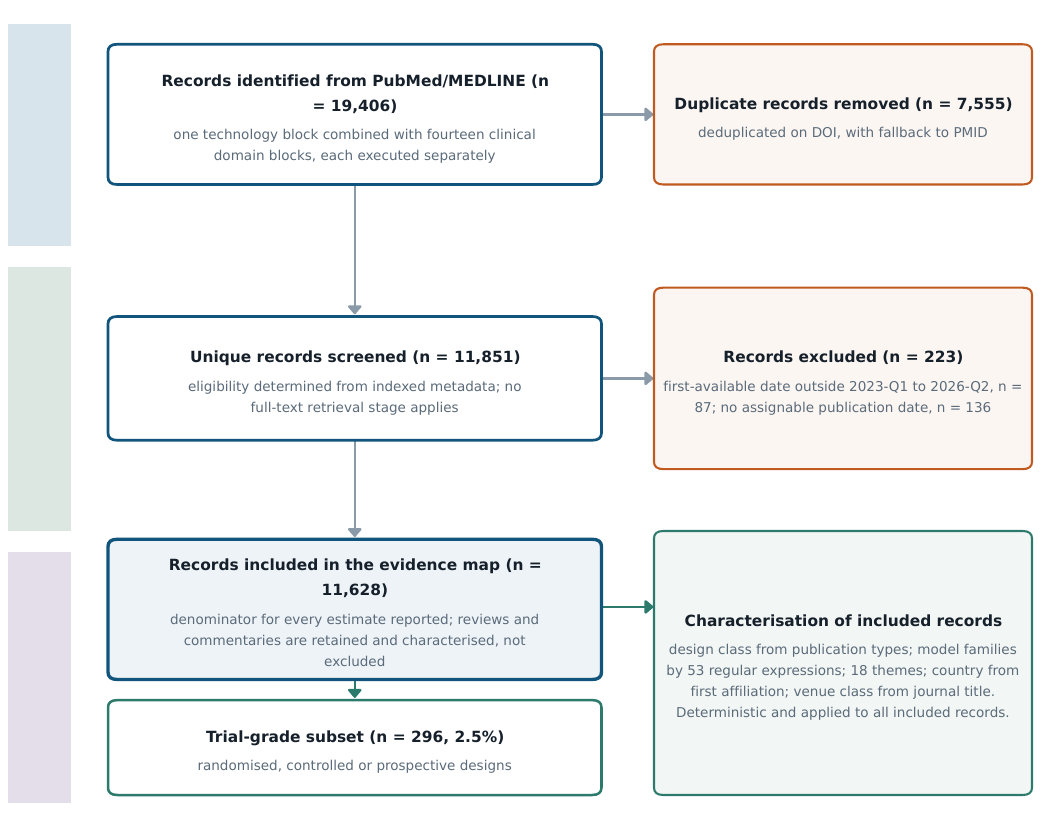}
\caption{\textbf{Study flow.} PRISMA 2020 flow adapted for an evidence map
built from indexed metadata: eligibility is determined from the record itself,
so the stages for reports sought and reports assessed for eligibility do not
apply. Exclusions, shown at right, are by publication date alone; no record was
excluded on judgement. The 11,628 included records are the denominator for
every estimate reported. Reviews and commentaries are retained and
characterised alongside empirical reports rather than excluded, because study
design is one of the quantities under analysis. Characterisation is summarised
at lower right and set out in full in Fig.~\ref{fig:workflow}; it is applied
identically to every included record. The trial-grade subset comprises
randomised, controlled and prospective designs, and is the basis for every
statement about evidence able to support inference on clinical use.}
\label{fig:flow}
\end{figure}

\begin{figure}[H]\centering
\includegraphics[width=\textwidth]{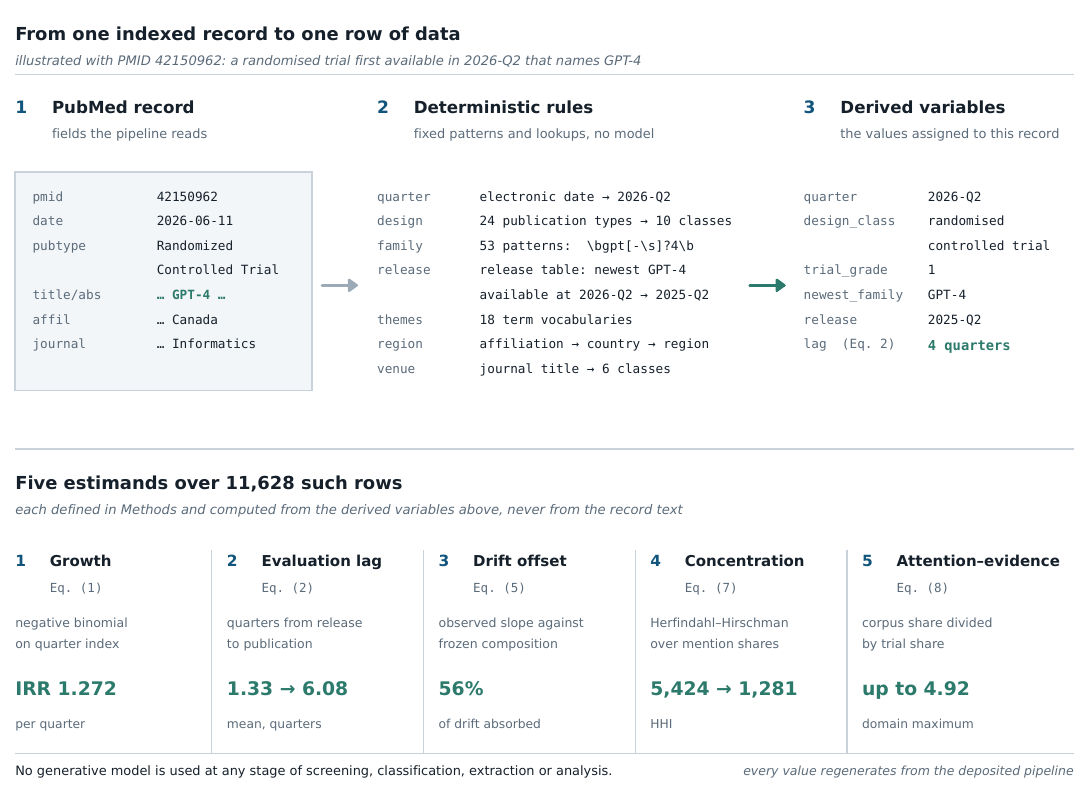}
\caption{\textbf{Study workflow.} The question, the search, the
characterisation rules applied to every included record, and the five analyses
those rules support. Each rule is a fixed pattern or lookup table applied
identically to all records; no generative model is used at any stage of
screening, classification, extraction or analysis. Values beneath each
analysis box give the corresponding headline estimate.}
\label{fig:workflow}
\end{figure}

Principal estimates are collected in Table~\ref{tab:principal}. The search
returned 19,406 records across fifteen query blocks, reducing to
11,851 unique records after deduplication on DOI with fallback to PMID. Of
these, 87 carried a first-available date outside the analysis window and 136
carried no date from which a quarter could be assigned, leaving 11,628 records
in the evidence map (Fig.~\ref{fig:flow}). Reviews and commentaries were
retained and characterised rather than excluded, since study design is one of
the quantities under analysis. The workflow from search to analysis is set out
in Fig.~\ref{fig:workflow}.

\begin{figure}[H]\centering
\includegraphics[width=\textwidth]{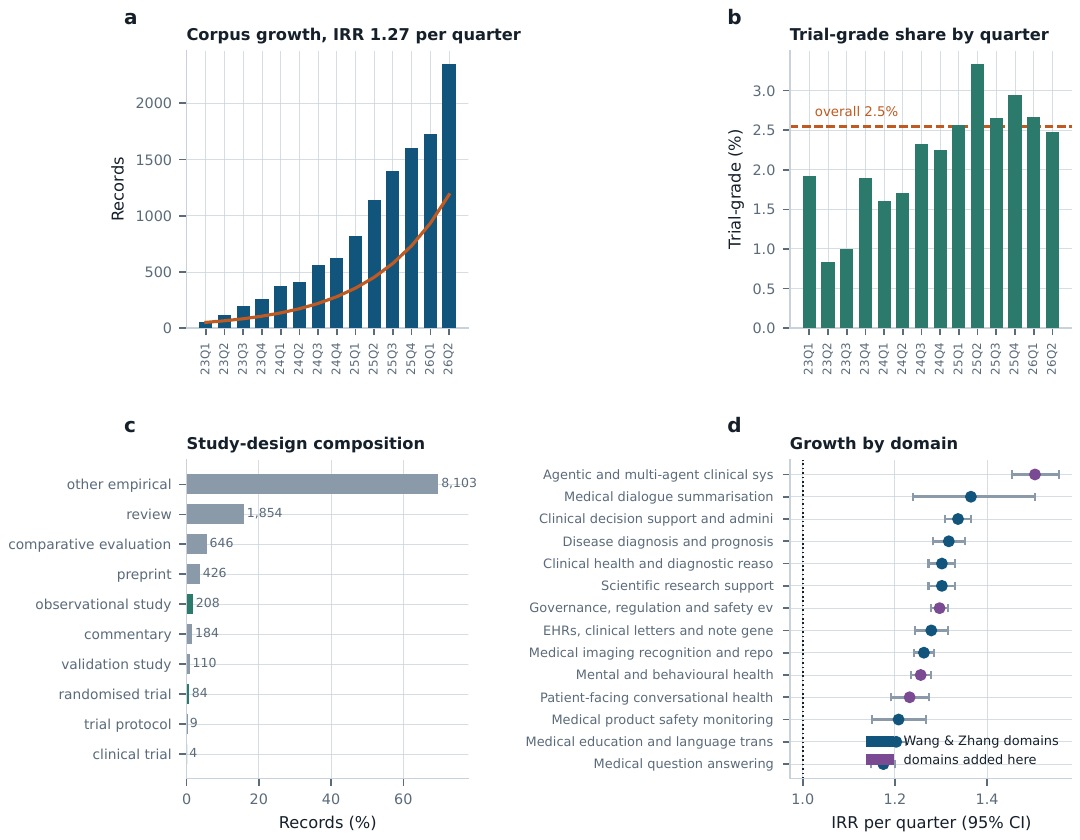}
\caption{\textbf{Growth and composition of the evidence base.}
\textbf{a}, Quarterly record counts with fitted negative binomial trend;
the fitted incidence rate ratio is 1.272 per quarter (95\% CI 1.257--1.289),
from 52 records in 2023-Q1 to 2,346 in 2026-Q2.
\textbf{b}, Trial-grade share by quarter; dashed line is the corpus mean.
\textbf{c}, Study-design composition; trial-grade designs highlighted.
\textbf{d}, Growth by domain, incidence rate ratio per quarter with 95\%
confidence intervals, coloured by whether the domain derives from the earlier
taxonomy or was added here.}
\label{fig:growth}
\end{figure}

Output grew at an incidence rate ratio of 1.272 per quarter (95\% CI
1.257--1.289), from 52 records in 2023-Q1 to 2,346 in 2026-Q2, a 45-fold
increase (Fig.~\ref{fig:growth}a). Growth was not uniform across domains
(Fig.~\ref{fig:growth}d). Agentic and multi-agent clinical systems grew
fastest (IRR 1.504, 95\% CI 1.454--1.556), followed by medical dialogue
summarisation (1.365, 1.240--1.504) and clinical decision support (1.337,
1.308--1.366). Medical question answering grew slowest (1.175, 1.149--1.201),
consistent with a maturing benchmark literature.

Study designs able to support inference about clinical use were scarce. In
total 296 records, 2.5\% of the corpus, met the trial-grade definition of a
randomised, controlled or prospective design; randomised controlled trials
accounted for 84 records (0.7\%), clinical trials for 4 and observational
studies for 208 (Fig.~\ref{fig:growth}c). The trial-grade share did not fall
over the window but neither did it rise materially in its second half: from
1.9\% in 2023-Q1 it reached a maximum of 3.3\% in 2025-Q2 and stood at 2.5\%
in 2026-Q2, and the last five quarters show no upward movement
(Fig.~\ref{fig:growth}b). A binomial model of the whole trajectory gives a
logit slope of 0.061 per quarter ($P=0.013$), but the value in the opening
quarter rests on a single trial-grade record and the recent plateau makes
extrapolation from this slope unsafe; we report it as a description of the
window rather than as a forecast.

\subsection*{The evaluation gap}

We defined evaluation lag as the number of quarters between the release of the
newest model family a study names and the study's own publication quarter. Lag
could be computed for 5,775 records; four named a model released after their
own publication date and were excluded.

\begin{figure}[H]\centering
\includegraphics[width=\textwidth]{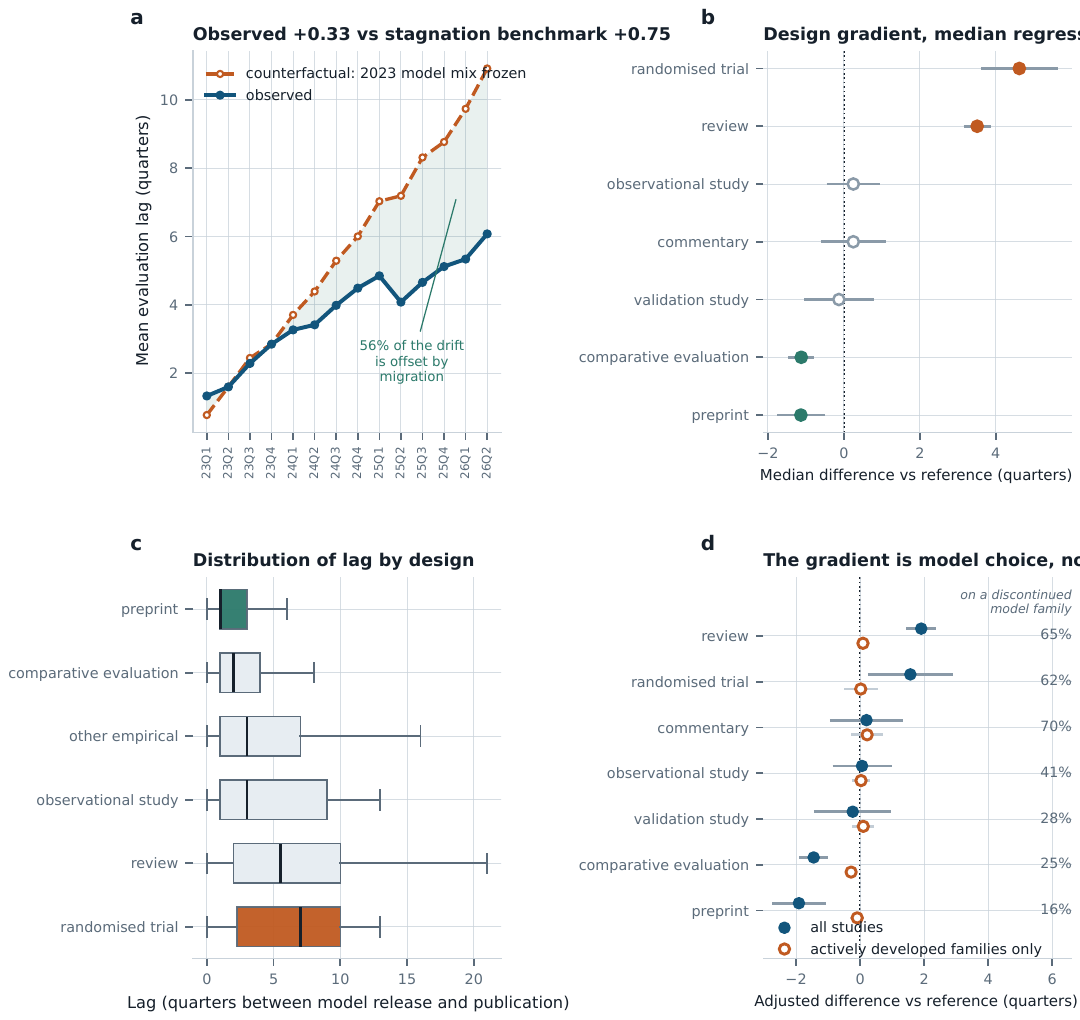}
\caption{\textbf{The evaluation gap.} \textbf{a}, Mean evaluation lag by
quarter (solid) against a counterfactual in which the composition of model
families is held at its 2023 value and only within-family lag evolves
(dashed). Because a family that receives no further releases ages at one
quarter per quarter, the dashed line is the drift the literature would
accumulate without changing what it studies; the shaded area is the part of
that drift offset by migration to newer systems. \textbf{b}, Design
differences in lag from median regression adjusted for publication quarter,
relative to other empirical designs; filled markers indicate $P<0.05$.
\textbf{c}, Distribution of lag by design; boxes show median and interquartile
range, whiskers 1.5\,IQR, outliers omitted. \textbf{d}, Adjusted design
differences among all studies (filled) and among studies naming a model family
still receiving releases (open); labels give the share of each design
evaluating a discontinued family. Median lag by individual model family is
given in Supplementary Fig.~S1.}
\label{fig:gap}
\end{figure}

Mean lag rose from 1.33 quarters in 2023-Q1 to 6.08 in 2026-Q2, that is from
approximately four to eighteen months (Fig.~\ref{fig:gap}a). Interpreting that
rise requires a benchmark. A model family that receives no further releases
ages at exactly one quarter per quarter, so lag would widen even if no
researcher changed what they studied, and a null of zero is uninformative.
Holding the composition of model families at its 2023 value and allowing only
within-family lag to vary gives a counterfactual slope of 0.753 quarters per
quarter, under which mean lag would have reached 10.92 quarters by 2026-Q2
rather than the observed 6.08. Against that benchmark the observed slope of
0.329 represents a migration to newer systems that offsets 56.2\% of the drift
the literature would otherwise accumulate (95\% confidence interval 49.9 to
65.2, 2,000 bootstrap resamples; 64.7\% taking the first quarter alone as
baseline). The same decomposition by model-family fixed effects gives a
within-family slope of 0.654 and a compositional component of $-$0.325. The
literature is refreshing its model base, but at little more than half the rate
required to hold the gap constant.

The widening was not an artefact of a residual literature on obsolete systems.
Encoder-era models---BERT and its biomedical derivatives, T5, GPT-3, BioGPT and
GatorTron---accounted for 198 records, 3.4\% of those with a computable lag,
and excluding them left the slope unchanged at 0.299 (95\% CI 0.268 to 0.330).
Restricting to the 3,468 records naming a family still receiving releases in
2024 or later reduced the slope to 0.237 (0.222 to 0.252) without removing it.
Records naming only discontinued families widened at 1.062 quarters per quarter
(1.013 to 1.110), indistinguishable from the mechanical rate of one, which
validates the decomposition. Under a pessimistic sensitivity analysis, in which
every study is assumed to have used the earliest release of each family it
names, the slope rose to 0.673 (0.643 to 0.703).

Lag differed systematically by study design. Because the distribution is
strongly right-skewed we estimated design differences by median regression on
publication quarter, taking other empirical designs as reference. Randomised
controlled trials evaluated models 4.62 quarters older than reference (95\% CI
3.62 to 5.63, $P=3.4\times10^{-19}$) and reviews 3.51 quarters older (3.15 to
3.87, $P=6.5\times10^{-80}$), whereas preprints evaluated models 1.13 quarters
more recent ($-$1.75 to $-$0.50, $P=4.3\times10^{-4}$) and comparative
evaluations 1.12 quarters more recent ($-$1.46 to $-$0.79,
$P=9.4\times10^{-11}$) (Fig.~\ref{fig:gap}b). Observational studies did not
differ from reference (0.25, $-$0.45 to 0.95, $P=0.48$), so the effect was
specific to randomised designs rather than to prospective designs in general.
Ordinary least squares on the same specification reproduced the ordering with
smaller coefficients; after Holm correction across the seven design contrasts
the estimates for preprints, comparative evaluations and reviews remained
significant and that for randomised trials did not (adjusted difference 1.57
quarters, $P=0.021$, Holm-adjusted $P=0.084$). An alternative
operationalisation, counting model generations behind the frontier rather than
quarters since release, placed randomised controlled trials furthest behind of
any empirical design (2.81 generations, $n=58$) and preprints among the closest
(1.73), and its ranking of designs agreed with the quarters-based estimate
(Spearman $\rho=0.72$).

The gradient reflects which systems each design evaluates rather than how long
each design takes to complete. Among records naming an actively developed model
family, no design differed materially from reference: randomised trials by 0.02
quarters (95\% CI $-$0.50 to 0.55, $P=0.93$, $n=22$), preprints by $-$0.09
($-$0.31 to 0.13, $P=0.43$) and reviews by 0.09 ($-$0.10 to 0.28, $P=0.34$);
only comparative evaluations retained a small difference ($-$0.28, $-$0.40 to
$-$0.15) (Fig.~\ref{fig:gap}d). What separated the designs was the probability
of studying a discontinued family at all: 62.1\% of randomised trials and
64.8\% of reviews, against 24.9\% of comparative evaluations and 15.6\% of
preprints. A single family accounted for much of this, with 2,070 records
naming GPT-3.5 or ChatGPT at a median lag of eight quarters, against one
quarter for GPT-4, Claude, Gemini and the OpenAI o-series (Supplementary
Fig.~S1). The subgroup of randomised trials naming a current model is small, so
this establishes that the gradient operates through model selection rather than
demonstrating the absence of any residual timeline effect. It is, however,
largely insulated from detection error: of its 22 records, 11 name GPT-4 and 4
name DeepSeek, neither of which can be matched by any clinical term, and the
measured flag rates imply fewer than one spurious record in the subgroup.

In the raw distribution the separation was wider than the adjusted medians
suggest. The median randomised trial evaluated a model released seven quarters
(21 months) earlier, interquartile range 2 to 10 quarters; the median preprint
evaluated one released a single quarter (3 months) earlier, interquartile range
1 to 3 (Fig.~\ref{fig:gap}c). Lag is defined only for records naming a model
family in the title or abstract, and coverage varied by design, from 87.0\% of
comparative evaluations and 69.0\% of randomised trials to 36.2\% of preprints
and 27.0\% of reviews. Estimates for designs with low coverage therefore
describe the subset that names a model, which is likely to be the more
completely reported subset.

\subsection*{Specification of model versions}

\begin{figure}[H]\centering
\includegraphics[width=0.86\textwidth]{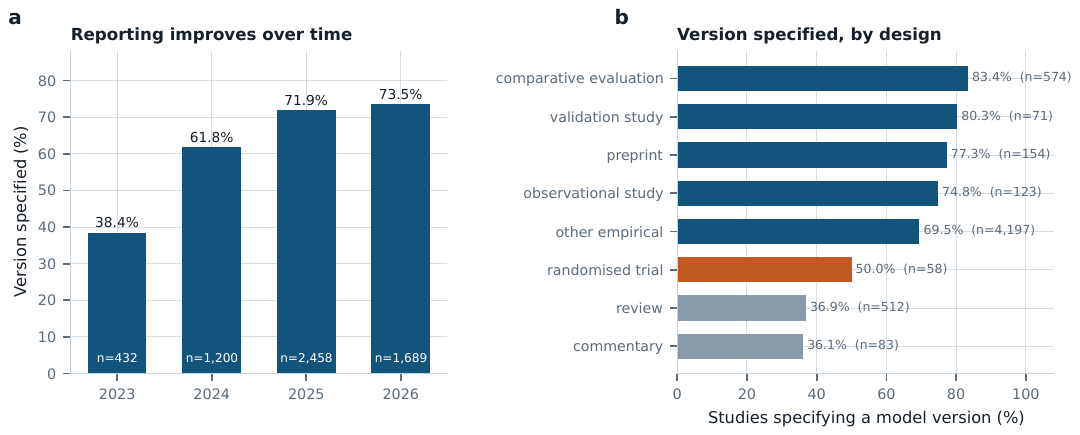}
\caption{\textbf{Specification of model versions.} \textbf{a}, Share of
records naming a model that also specify a version, access date or decoding
parameter, by year. \textbf{b}, The same share by study design, restricted to
designs with at least 40 records.}
\label{fig:version}
\end{figure}

Of records naming at least one model family, 67.8\% specified a version, an
access date, a decoding parameter or an equivalent identifier sufficient to
determine which system was evaluated. Specification improved over the window,
from 38.4\% of such records in 2023 to 73.5\% in 2026
(Fig.~\ref{fig:version}a).

Specification differed markedly by design (Fig.~\ref{fig:version}b).
Comparative evaluations specified a version in 83.4\% of cases ($n=574$) and
validation studies in 80.3\% ($n=71$), whereas randomised controlled trials
did so in 50.0\% ($n=58$), the lowest of any empirical design and below
preprints (77.3\%, $n=154$) and observational studies (74.8\%, $n=123$).
Reviews and commentaries specified least often (36.9\% and 36.1\%). Half of
the randomised evidence in this corpus therefore cannot be attributed to a
determinate model version.

\subsection*{A diversifying but not specialising model landscape}

\begin{figure}[H]\centering
\includegraphics[width=\textwidth]{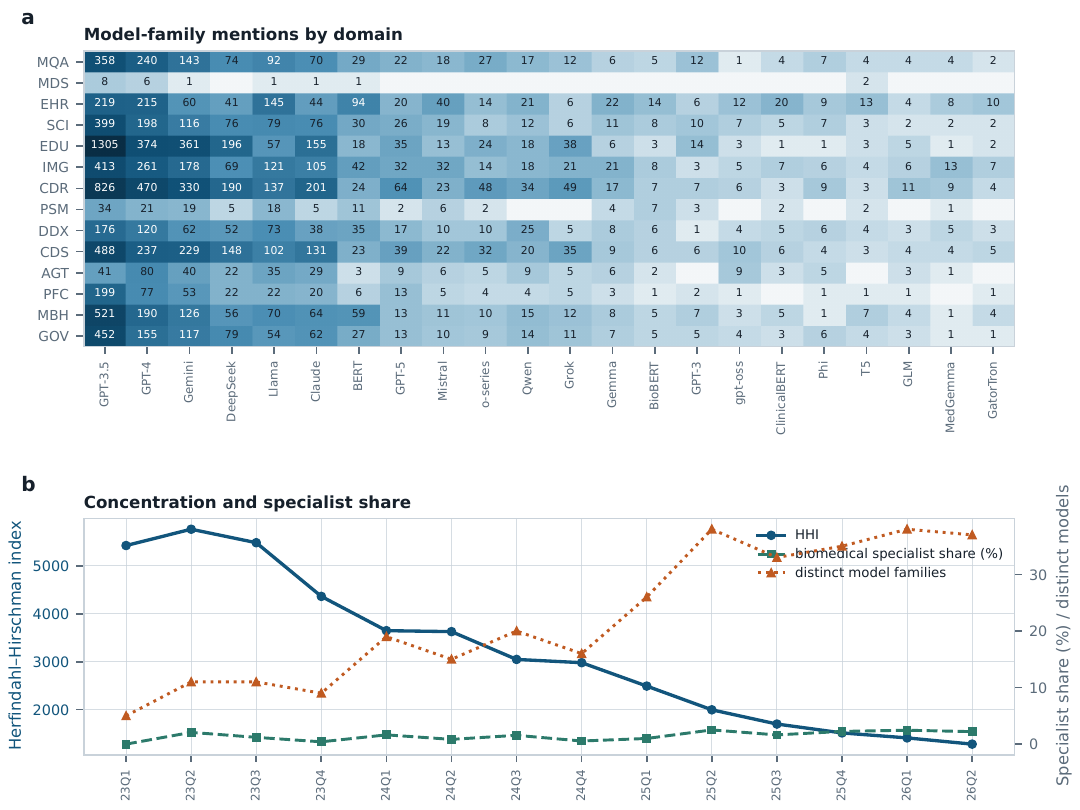}
\caption{\textbf{The model landscape.} \textbf{a}, Model-family mentions by
clinical domain on a logarithmic colour scale; a record naming several
families contributes several mentions. \textbf{b}, Concentration over time:
Herfindahl--Hirschman index, number of distinct families named, and share of
mentions held by purpose-built biomedical models.}
\label{fig:landscape}
\end{figure}

Model use was heavily concentrated at the start of the window and became
markedly less so. The Herfindahl--Hirschman index of model-family mentions
fell from 5,424 in 2023-Q1 to 1,281 in 2026-Q2, a decline of 76\%, while the
number of distinct families named in a quarter rose from 5 to 37 and the share
held by the most-named family fell from 72.0\% to 25.3\%
(Fig.~\ref{fig:landscape}b).

Diversification occurred almost entirely among general-purpose systems. Models
built specifically for biomedical use---including Med-PaLM, MedGemma,
Meditron, OpenBioLLM, BioMistral and HuatuoGPT\cite{Sellergren2025}---accounted for 0.0\% of
mentions in 2023-Q1 and 2.21\% in 2026-Q2, and did not exceed 2.6\% in any
quarter. Across the corpus their share was 1.63\% of 15,506 mentions. Model
use was not independent of clinical domain ($\chi^2=2{,}386$, $df=676$,
$P<0.001$; Cram\'er's $V=0.109$), but the dominant family was the same in
thirteen of fourteen domains; only agentic and multi-agent systems\cite{Qiu2024} had a
different leader (Fig.~\ref{fig:landscape}a).

\subsection*{Where attention and evidence diverge}

\begin{figure}[H]\centering
\includegraphics[width=\textwidth]{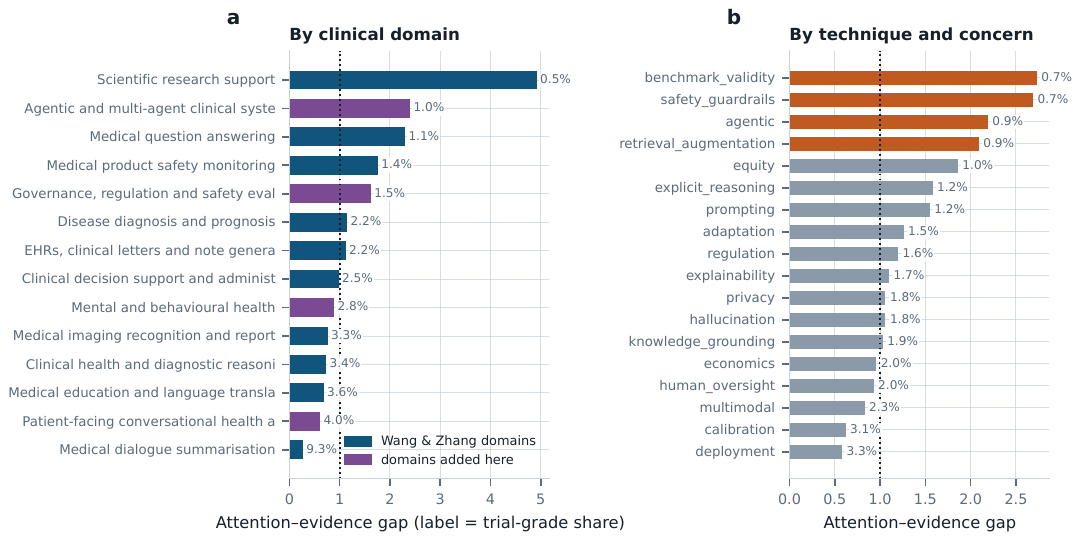}
\caption{\textbf{Attention--evidence gap.} Corpus share divided by trial-grade
share, for clinical domains (\textbf{a}) and for techniques and concerns
(\textbf{b}). Values above one indicate publication volume exceeding
trial-grade representation. Labels give each category's trial-grade share.}
\label{fig:aeg}
\end{figure}

We expressed the mismatch between publication volume and trial-grade evidence
as an attention--evidence gap: a domain's share of the corpus divided by its
share of trial-grade records (Fig.~\ref{fig:aeg}a). Scientific research
support showed the largest gap (4.92; 1,390 records, 7 trial-grade),
followed by agentic and multi-agent clinical systems (2.40; 678 records, 7
trial-grade) and medical question answering (2.30; 835 records, 9
trial-grade). At the other extreme, medical dialogue summarisation had the
smallest gap (0.27) and the highest trial-grade share of any domain (9.26\%),
despite being the smallest domain in the corpus (54 records).

The same measure applied to techniques and concerns
(Fig.~\ref{fig:aeg}b) placed benchmark validity (2.73; 0.7\% trial-grade) and
safety guardrails (2.69; 0.71\%) at the top: the two themes most directly
concerned with establishing trustworthiness had the least trial-grade evidence
behind them. Retrieval augmentation (2.09) and agentic behaviour (2.19) were
also markedly evidence-poor relative to attention.

\subsection*{Two literatures}

\begin{figure}[H]\centering
\includegraphics[width=\textwidth]{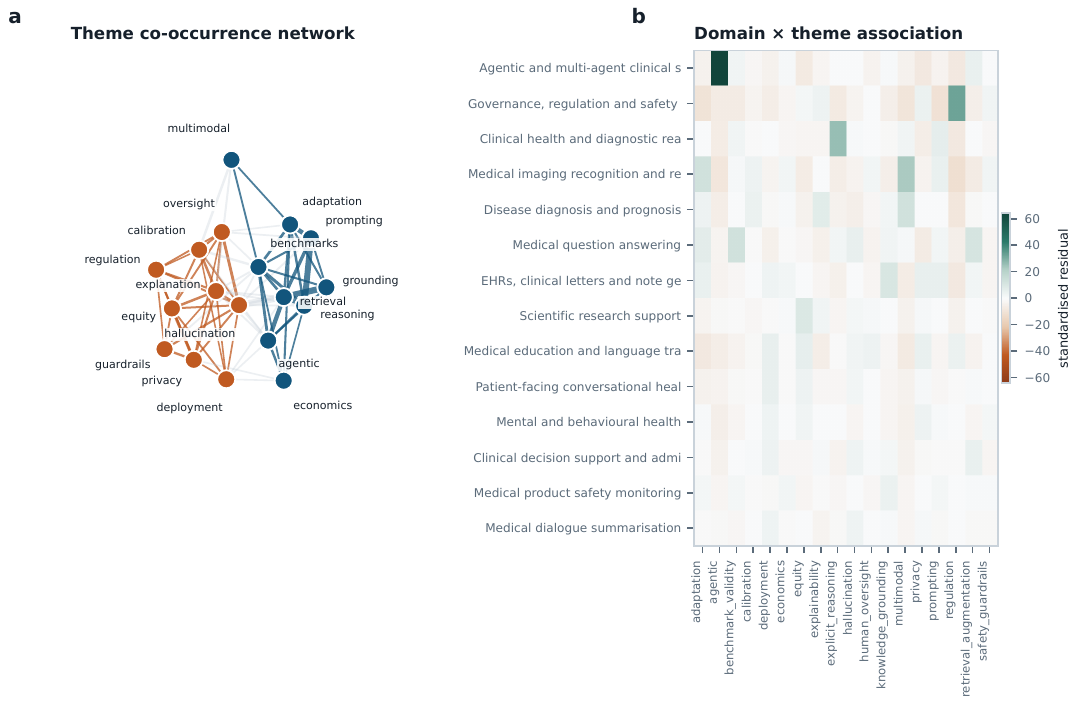}
\caption{\textbf{Two literatures.} \textbf{a}, Theme co-occurrence network,
edges drawn where lift exceeds 1.25 with at least 30 co-occurrences; colour
denotes the community identified by modularity maximisation; labels are
abbreviated. \textbf{b}, Standardised Pearson residuals for the association
between clinical domain and theme.}
\label{fig:two}
\end{figure}

Themes did not co-occur at random. Modularity analysis of the theme
co-occurrence network resolved two communities of nine themes each
(Fig.~\ref{fig:two}a). The first comprised agentic behaviour, retrieval
augmentation, knowledge grounding, explicit reasoning, prompting, adaptation,
multimodality, benchmark validity and economics. The second comprised
hallucination, calibration, safety guardrails, human oversight, explainability,
privacy, equity, regulation and deployment. The strongest single
association bridged the two communities: hallucination with retrieval
augmentation (lift 3.95), retrieval being the standard technical response to
the failure mode. Every other association of comparable strength lay within
the technical community---knowledge grounding with retrieval augmentation
(3.78), explicit reasoning with prompting (3.62), adaptation with prompting
(2.78)---so the two literatures meet at one point rather than along a front.

Domain and theme were strongly associated ($\chi^2=10{,}110$, $df=221$, $P<0.001$;
Cram\'er's $V=0.139$; Fig.~\ref{fig:two}b). The largest standardised residuals
fell on the two domains absent from the earlier taxonomy: agentic systems with
the agentic theme (64.3) and governance with the regulation theme (32.3),
followed by diagnostic reasoning with explicit reasoning (25.8) and imaging
with multimodality (22.9). Negative residuals were concentrated where the two
communities meet, such as imaging with regulation ($-$11.3) and governance
with prompting ($-$10.4).

The four domains added to the earlier taxonomy accounted for 27.1\% of
domain-record assignments, with a trial-grade share (2.1\%) similar to that of
the ten inherited domains (2.6\%).

\subsection*{Distribution across regions, venues and journals}

\begin{figure}[H]\centering
\includegraphics[width=\textwidth]{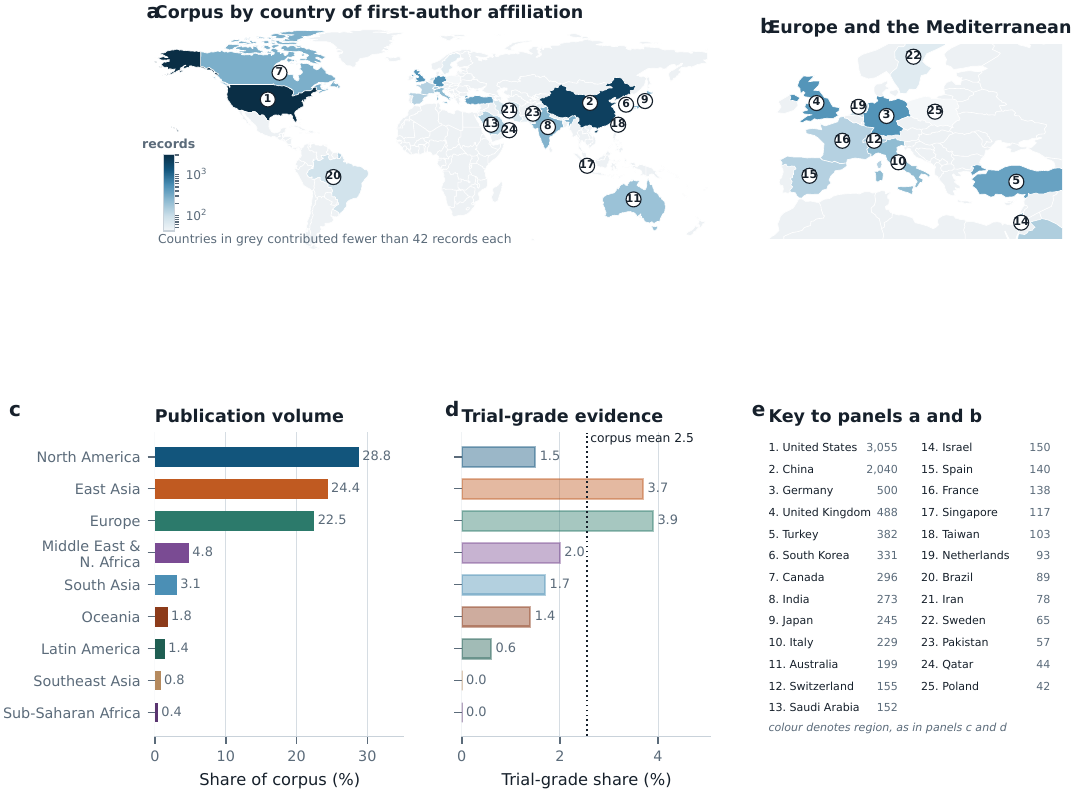}
\caption{\textbf{Geographic distribution of the corpus and of trial-grade
evidence.} \textbf{a}, Records by country of first-author affiliation on a
logarithmic colour scale; numbers key to panel \textbf{e}, and countries in
grey contributed fewer records than the twenty-fifth most productive country.
\textbf{b}, Europe and the Mediterranean, where country areas are too small to
label on the world projection. \textbf{c}, Share of the corpus by region.
\textbf{d}, Trial-grade share within each region on the same vertical
ordering; the dotted line is the corpus mean of 2.5\%. Among the three
highest-output regions the two panels run opposite to one another: North
America leads on volume and trails Europe and East Asia on trial-grade share.
The lower shares below them arise in regions contributing fewer than 400
records, where the estimate is unstable. \textbf{e}, Key to panels \textbf{a} and
\textbf{b} with record counts; colour denotes region.}
\label{fig:geo}
\end{figure}

Corresponding authorship was concentrated in North America (28.8\%), East Asia
(24.4\%) and Europe (22.5\%); 12.0\% of records could not be attributed to a
region from the first affiliation (Fig.~\ref{fig:geo}a,~c). The United States
contributed 3,055 records and China 2,040, together 44\% of the corpus
(Fig.~\ref{fig:geo}e). Among the three regions producing most of the
literature, trial-grade share ran opposite to volume: 3.9\% in Europe and
3.7\% in East Asia against 1.5\% in North America, which published most and
tested least of the three (Fig.~\ref{fig:geo}d). Lower shares occurred only in
regions whose output was too small to interpret, with no trial-grade records
among the 89 records from Southeast Asia or the 43 from sub-Saharan Africa.

Clinical and biomedical journals carried 46.2\% of the corpus at a trial-grade
share of 3.4\%. High-impact general medical journals carried 2.1\% at a
trial-grade share of 5.7\%, the highest of any venue class, whereas computer
science venues carried 3.1\% and preprint servers 3.7\%, neither containing a
single trial-grade record. The
literature was distributed across 2,304 journals, of which 32 (1.4\%) carried
the first third of records, consistent with a Bradford distribution with
successive multipliers of 7.6 and 8.4 (Supplementary Fig.~S2).

\section{Discussion}

Across 11,628 indexed reports on generative language models in healthcare, we
find an evidence base expanding at 1.27 times per quarter while the models it
evaluates recede further into the past. Mean evaluation lag widened from
approximately four months to eighteen over three and a half years. That
widening is not simply models growing older: the literature does migrate
toward newer systems, and the migration offsets a little over half of the
drift it would otherwise accumulate. It is the shortfall that matters, and the
shortfall survives every restriction we applied, including removal of the
entire encoder-era literature and restriction to families still under active
development.

The association with study design initially suggests research timelines.
Preprints, which have no peer review and the shortest interval between conduct
and appearance, evaluated the most recent models. Randomised controlled
trials, which require protocol development, approval, recruitment, follow-up
and peer review, evaluated models a median of 21 months old. Reviews were
furthest behind of all, a finding that applies to the present work and which we
address by depositing the pipeline so that the map can be regenerated rather
than repeated.

The stratified analysis does not support that explanation. Among studies
evaluating a model family still receiving releases, no design differed from any
other: randomised trials, preprints and reviews were statistically
indistinguishable. What distinguished randomised trials was not that they took
longer but that 62\% of them evaluated a system whose developer had stopped
shipping releases, against 16\% of preprints. Randomised evaluations of
artificial intelligence in clinical practice remain scarce in absolute
terms\cite{Han2024}, so the systems they select determine what the strongest
tier of evidence is about. On this evidence the distance
between rigorous evidence and current technology is not principally a
consequence of how long rigorous work takes. It is a consequence of what
rigorous work chooses to evaluate, and that is a modifiable feature of study
design rather than an irreducible cost of doing careful research. This matters most where the evidence is meant to
inform deployment rather than benchmark performance\cite{Wekenborg2025}, and
where the quantity of interest is the performance of a clinician working with
the system rather than of the system alone\cite{Vaccaro2024}.
Trials could specify the newest available model at the point of protocol
registration and adapt, or could be designed so that the model is a
replaceable component rather than the object of study.

The reproducibility finding compounds this. Randomised trials specified the
model version they evaluated in half of cases, the lowest rate of any
empirical design and well below comparative evaluations at 83.4\%. A trial
that reports neither the version nor the access date of a proprietary system
cannot be reproduced even in principle, because the system it evaluated no
longer exists in that form. The highest tier of evidence in this literature is
thus both the least current and the least attributable. This is a specific,
addressable reporting failure: CONSORT-AI\cite{Liu2020},
DECIDE-AI\cite{Vasey2022}, TRIPOD-LLM\cite{Gallifant2025} and the CHART
statement\cite{CHART2025} already require model provenance, and enforcement at the point of submission would be
sufficient.

The concentration analysis revises a common characterisation of this field.
Model use was extremely concentrated in early 2023 and is no longer: the
Herfindahl--Hirschman index fell by 76\%, and the number of families in use
rose sevenfold. Descriptions of the field as dominated by a handful of
proprietary systems are becoming outdated; open-weight systems now undergo
independent clinical benchmarking\cite{Sandmann2025}. What has not changed is the
position of purpose-built biomedical models, which have never exceeded 2.6\%
of mentions in any quarter despite substantial development activity. The
evidence base for medical language models is, empirically, an evidence base
about general-purpose systems adapted to medical tasks. Whether that reflects
access, cost, capability or convention cannot be determined from bibliometric
data, but the consequence is that findings inherit the behaviour and release
schedules of systems developed outside medicine.

The attention--evidence gap identifies where these problems concentrate.
Scientific research support, the largest gap at 4.92, contains 1,390 reports
and seven trial-grade studies. Agentic and multi-agent systems, the
fastest-growing domain in the corpus, contains 678 reports and seven. That
these are also the domains where autonomy is greatest is uncomfortable: the
applications delegating most to the model are those with least prospective
evaluation. At the theme level, benchmark validity and safety guardrails rank
first and second, so the questions of whether evaluations measure the right
thing and whether deployed systems are constrained appropriately are the
questions least often addressed with trial-grade designs.

The modularity result suggests why this may persist. The literature separates
into a technical community concerned with capability---agentic behaviour,
retrieval, reasoning, prompting, adaptation---and a governance community
concerned with hallucination, oversight, regulation, equity and privacy. These
clusters share few papers. Work that advances capability rarely carries safety
analysis in the same report, and work on safety rarely advances capability.
The two literatures address the same systems with little overlap in
authorship or framing, and the domains most affected are those where one
community is nearly absent.

Several limitations qualify these findings. PubMed indexes clinical and
biomedical literature comprehensively but computer science venues only
partially, so work published at machine learning conferences is
under-represented; the observation that computer science venues contain no
trial-grade records should be read in that light. Model detection relies on
regular expressions over titles and abstracts, so a study evaluating a model
named only in its methods section is not captured, and family-level resolution
means a study naming GPT-4 is assigned the most recent GPT-4 release available
at its publication date, which is deliberately conservative with respect to
the lag hypothesis. Detection is also exposed to collisions between model
names and terms that occur independently in the clinical literature, chiefly
trial acronyms and anatomical abbreviations; a precision audit of the eleven
most exposed families flagged 9.9\% of their matches, or 2.0\% of the corpus,
and the direction of any residual error differs by family, inflating the
long-lag tail where a discontinued family is affected and deflating it where an
active one is. The families that dominate the corpus are unaffected, and the
subgroup carrying the design finding is expected to contain fewer than one
spurious record. Evaluation lag is consequently defined for only half the
corpus, and coverage is uneven across designs, so lag estimates for preprints
and reviews describe the minority of those records that name a model
explicitly. The finding that the design gradient disappears among actively
developed models rests on 22 randomised trials and is powered to detect only
large residual differences; it establishes that model selection accounts for
the gradient, not that no timeline effect exists. Classification of a family as
actively developed depends on the release table, which is the component of the
analysis most open to challenge. Release dates were assigned by hand and the full table is deposited. Design
classification depends on PubMed publication types, which are applied by
indexers and can lag or omit; this most plausibly under-counts trial-grade
records. Regional attribution used first-author affiliation and failed for
19.0\% of records. Records in the final quarter are subject to indexing lag,
and all trend estimates are reported with and without it. Finally, the
analysis is bibliometric: it describes what the literature evaluates and
reports, not whether the systems evaluated are safe or effective.

The practical implications are narrow and actionable. Trials of generative
systems should report the model version, access date and decoding parameters
as a minimum, and journals can enforce this at submission. Trial designs that
treat the model as a replaceable component, rather than as the object of
study, would allow findings to outlast a release cycle. Where a domain shows a
large attention--evidence gap, that is the argument for prospective work
rather than further benchmarking. And because the gap widens continuously,
evidence syntheses in this area should be built to be regenerated; the
pipeline used here is deposited for that purpose.

\section{Methods}

\subsection*{Design and reporting}

This is a meta-research study: a reproducible evidence map of the indexed
literature, in which the unit of analysis is the published report rather than
the patient or the clinical outcome. It does not synthesise treatment effects
and does not assess risk of bias in individual studies, neither of which is
applicable to the questions asked. Reporting follows PRISMA 2020 where
applicable to search and selection\cite{Page2021}; the completed checklist is
provided as Supplementary Table~S1. The study was registered on OSF Registries
after analysis was complete; the record states this explicitly and lists all
deviations from the original analysis plan. PROSPERO was not used because it
does not accept meta-research. The registration, the complete search strings
(Supplementary Table~S2), the code and all data underlying the figures are
deposited (see Data availability and Code availability).

\subsection*{Search}

PubMed/MEDLINE was searched through the NCBI E-utilities interface for records
with a first-available publication date between 1 January 2023 and 30 June
2026. The search combined one technology block with fourteen clinical domain
blocks, each executed separately and then deduplicated on DOI with fallback to
PMID. The technology block covered generic terminology (large language model,
generative artificial intelligence, foundation model, vision--language model,
reasoning model) and named systems (GPT-3 through GPT-5, ChatGPT, Gemini,
Claude, Llama, DeepSeek, Qwen, Mistral, Med-PaLM, MedGemma, Meditron,
OpenBioLLM, HuatuoGPT). Complete executable strings for every block are given in Supplementary
Table~S2, which is generated directly from the vocabulary module the
harvester imports so that the reported and executed searches cannot diverge;
per-block yields and execution timestamps are written by the harvesting
script.

Ten domain blocks follow the application taxonomy established by Wang and
Zhang for this literature\cite{Wang2024}, with terminology updated. Four
further domains were added to cover application areas that postdate their
January 2024 search: agentic and multi-agent clinical systems; patient-facing
conversational health and triage; mental and behavioural health; and
governance, regulation and safety evaluation. Records may be assigned to more
than one domain.

Only PubMed was searched. The decision reflects the clinical framing of the
question and the requirement that design classification be derived from
indexed metadata rather than manual coding; it is a limitation, addressed in
the Discussion.

\subsection*{Record characterisation}

All characterisation is programmatic and deterministic. Design class was
derived from PubMed publication types using a fixed priority ordering, mapping
to ten analysis classes; records typed as randomised controlled trial,
pragmatic clinical trial, equivalence trial, clinical trial or observational
study were designated trial-grade. Publication quarter was taken from the
electronic article date where present and from the journal issue date
otherwise. Country was resolved from the first author affiliation against a
188-entry lookup and mapped to nine regions. Journal titles were mapped to six
venue classes by regular expression. Model families were detected by 53
regular expressions over title and abstract, disambiguating families that
share substrings (for example MedGemma from Gemma, ClinicalBERT from BERT, and
GPT-4o from both GPT-4 and the o-series). Eighteen cross-cutting themes
covering techniques and concerns were detected in the same way. The complete
vocabulary is deposited as a single module and was fixed before analysis.

\subsection*{Analysis window}

Records whose first-available date fell outside 2023-Q1 to 2026-Q2 were
excluded from all estimates ($n=87$). Because indexing is incomplete for the
most recent quarter, every trend estimate is reported with and without the
final quarter.

\subsection*{Statistical analysis}

\paragraph{Growth.} Let $y_t$ be the number of records first available in
quarter $t=0,\dots,T$. Counts were modelled by negative binomial regression,
\begin{equation}
y_t \sim \mathrm{NB}(\mu_t,\alpha), \qquad
\log \mu_t = \beta_0 + \beta_1 t ,
\label{eq:nb}
\end{equation}
with the incidence rate ratio per quarter $\mathrm{IRR}=\exp(\beta_1)$ and its
95\% confidence interval $\exp(\beta_1 \pm 1.96\,\mathrm{SE}(\beta_1))$. The
dispersion parameter $\alpha$ was estimated by auxiliary regression on Poisson
fitted values. Monotonic trend was tested by the Mann--Kendall statistic
\cite{Mann1945} and structural breaks located by the Pettitt test
\cite{Pettitt1979}. Every trend is reported with and without the final
quarter, which is subject to indexing lag.

\paragraph{Evaluation lag.} Let record $i$ be published in quarter $q_i$ and
name the set of datable model families $\mathcal{F}_i$. Writing $r_f(q)$ for
the release quarter of the most recent release of family $f$ available at
quarter $q$, evaluation lag is
\begin{equation}
\lambda_i \;=\; q_i - \max_{f \in \mathcal{F}_i} r_f(q_i) .
\label{eq:lag}
\end{equation}
Taking the most recent release of each family, and the maximum over families
named, are both charitable to the study and therefore conservative with
respect to the hypothesis that lag is large. A sensitivity analysis replaces
$r_f(q_i)$ with the earliest release of $f$. Records with $\lambda_i<0$ were
flagged and excluded ($n=4$). The release table is deposited in full.

\paragraph{Benchmarking the lag trend.} A family that receives no further
releases satisfies $r_f(q)=r_f$ for all subsequent $q$, so its contribution to
$\lambda$ grows at exactly one quarter per quarter. Mean lag therefore rises
under unchanged research behaviour, and a null of zero slope is uninformative.
Decomposing mean lag in quarter $t$ over families,
\begin{equation}
L_t \;=\; \sum_{f} w_{ft}\,\ell_{ft},
\label{eq:decomp}
\end{equation}
where $w_{ft}$ is the share of records in quarter $t$ whose newest named
family is $f$ and $\ell_{ft}$ the mean lag among them, we construct a
counterfactual series that holds composition at its baseline value $w_f^{0}$
and lets only within-family lag evolve,
\begin{equation}
\tilde{L}_t \;=\;
\frac{\sum_{f} w_f^{0}\,\ell_{ft}\,\mathbb{1}\{\ell_{ft} \text{ observed}\}}
     {\sum_{f} w_f^{0}\,\mathbb{1}\{\ell_{ft} \text{ observed}\}} .
\label{eq:cf}
\end{equation}
Writing $\beta$ and $\tilde{\beta}$ for the ordinary least squares slopes of
$L_t$ and $\tilde{L}_t$ on $t$, the drift offset is
\begin{equation}
\Delta \;=\; 1 - \beta/\tilde{\beta},
\label{eq:offset}
\end{equation}
the proportion of the mechanical drift absorbed by migration to newer systems;
$\Delta=1$ corresponds to a literature that holds its lag constant and
$\Delta=0$ to one that never changes what it studies. This is a shift-share
decomposition in the sense of Oaxaca and Blinder \cite{Oaxaca1973,Blinder1973}.
$\Delta$ is reported with a percentile bootstrap confidence interval over
2{,}000 resamples of records \cite{Efron1993}, and the baseline period was
varied from the calendar year 2023 to the first quarter alone. The same
decomposition was performed by ordinary least squares with model-family fixed
effects, yielding within-family and compositional components. A family was
classified as actively developed if its most recent release fell in 2024-Q1 or
later and as discontinued otherwise; this is a property of the family, not of
any study, so conditioning on it does not condition on the outcome. Trend
estimates are reported for the whole sample, excluding encoder-era families,
and restricted to actively developed families.

\paragraph{Design differences.} Because the distribution of $\lambda$ is
strongly right-skewed, design differences were estimated by median regression
\cite{Koenker1978}, minimising
\begin{equation}
\sum_i \rho_{0.5}\!\left(\lambda_i - \mathbf{x}_i^{\top}\boldsymbol{\beta}\right),
\qquad \rho_\tau(u)=u\left(\tau - \mathbb{1}\{u<0\}\right),
\label{eq:qr}
\end{equation}
with $\mathbf{x}_i$ containing design-class indicators and publication quarter,
other empirical designs as reference, and classes of at least 15 records.
Ordinary least squares on the same specification is reported alongside, with
$P$ values adjusted across the seven design contrasts by the Holm procedure
\cite{Holm1979}. The analysis was repeated within the subset naming an
actively developed family. As an alternative operationalisation, lag was also
expressed as the number of model generations between the newest family a
record names and the generation at the frontier in its publication quarter,
using six generations dated by first release; the two measures are reported
separately and never combined. Because $\lambda$ is defined only for records
naming a model in the title or abstract, coverage is reported by design class.

\paragraph{Detection precision.} Model names can collide with terms that
occur independently in the clinical literature, chiefly trial acronyms and
anatomical abbreviations. We audited the eleven most exposed families by
flagging every match co-occurring with a study designator or an anatomical
context term. Across those families 232 of 2,347 matches were flagged (9.9\%),
highest for Gemini (11.9\%) and Claude (9.9\%); flagged matches are 2.0\% of
the corpus and an expected 2.5\% of records with a computable lag. Exposure is
concentrated in families whose names are short common words, and is nil for
the families that dominate the corpus, since no clinical term matches GPT-4,
GPT-3.5, DeepSeek or the OpenAI o-series. The audit is deposited.

\paragraph{Version specification.} Defined as the presence, in title or
abstract, of a version identifier, access date, decoding parameter or
equivalent, among records naming at least one model family. The requirement
follows CONSORT-AI, which lists algorithm version among the items a trial
report must specify \cite{Liu2020}, and TRIPOD-LLM \cite{Gallifant2025}.

\paragraph{Concentration.} With $s_{ft}$ the share of model-family mentions
held by family $f$ in quarter $t$ and $K_t$ the number of distinct families,
\begin{equation}
\mathrm{HHI}_t = 10{,}000 \sum_f s_{ft}^2 ,
\qquad
E_t = \frac{-\sum_f s_{ft}\ln s_{ft}}{\ln K_t} ,
\label{eq:hhi}
\end{equation}
the Herfindahl--Hirschman index \cite{Hirschman1964} and Shannon evenness
\cite{Shannon1948}, computed per quarter and per domain. Independence of model
use across domains was tested by $\chi^2$ with Cram\'er's $V=\sqrt{\chi^2 /
\left(N(\min(r,c)-1)\right)}$ \cite{Cramer1946}. Mentions are not records: a
record naming three families contributes three mentions.

\paragraph{Attention--evidence gap.} For category $c$ with $n_c$ records of
which $m_c$ are trial-grade, out of $N$ records and $M$ trial-grade records
overall,
\begin{equation}
\mathrm{AEG}_c \;=\; \frac{n_c/N}{m_c/M} ,
\label{eq:aeg}
\end{equation}
so values above one indicate publication volume exceeding trial-grade
representation.

\paragraph{Theme structure.} Co-occurrence of themes $a$ and $b$ was
quantified by lift against the product of marginals,
\begin{equation}
\mathrm{lift}(a,b) \;=\; \frac{n_{ab}\,N}{n_a\,n_b} ,
\label{eq:lift}
\end{equation}
and communities identified by greedy modularity maximisation
\cite{Clauset2004,Newman2006} on the graph retaining edges with
$\mathrm{lift}>1.25$ and at least 30 co-occurrences. Domain--theme association
was assessed by standardised Pearson residuals. Regional specialisation was
expressed as location quotients, and journal concentration summarised by
Bradford zones \cite{Bradford1934}. The trajectory of the trial-grade share
was modelled by binomial regression on quarter index; because the final five
quarters show no upward movement, the fitted slope is reported as a
description of the window and not extrapolated.

Analyses used Python 3.12 with pandas, NumPy, SciPy, statsmodels and networkx.

\section*{Data availability}

The derived data supporting every figure and every statistic reported here are
openly available at Zenodo, \texttt{https://doi.org/10.5281/zenodo.21671630}, under a
Creative Commons Attribution 4.0 licence. The deposit contains the complete set
of derived tables, the controlled vocabulary, the hand-assigned model release
table, the study registration, and a record index giving the PubMed identifier,
DOI, publication quarter, design class, country and model families detected for
every one of the 11,628 analysed records.

Bibliographic records retrieved from PubMed/MEDLINE are the property of the
National Library of Medicine and of the publishers of the indexed articles, and
titles and abstracts are therefore not redistributed. The record index and the
complete search strings (Supplementary Table~S2) allow the corpus to be
reconstituted exactly from PubMed by any user, and the harvesting script
performs that reconstruction unattended.

\section*{Code availability}

The complete pipeline, comprising harvesting, deterministic characterisation,
statistical analysis, figure generation and the automated layout and compliance
checks, is openly available at Zenodo, \texttt{https://doi.org/10.5281/zenodo.21671630},
under the MIT licence, and is developed at
\texttt{https://github.com/raadbintareaf/evaluation-gap-npj}. The archived
version corresponds to release \texttt{v1.0}.

Executing the deposited scripts in numerical order reconstitutes the corpus
from PubMed and regenerates every derived table and every statistic reported
here in a single pass. Two verification scripts are included and are run as part of that
pass: one confirms that the study flow closes arithmetically and that the
descriptive and modelled analyses share a single denominator, and the other
confirms that every figure satisfies the journal's layout requirements. No
generative model is used at any stage of screening, classification, extraction
or analysis.

\clearpage
\bibliography{refs}

\section*{Acknowledgements}

This research received no specific grant from any funding agency in the
public, commercial or not-for-profit sectors.

\section*{Author contributions}

R.B.T.\ conceived the study, designed the pipeline, implemented the analysis
and wrote the manuscript. M.A.-R.\ and S.L.\ contributed to the design of the
domain taxonomy and the classification vocabulary, verified the model release
table, and revised the manuscript. All authors approved the final version and
accept accountability for the work.

\section*{Competing interests}

The authors declare no competing interests.

\section*{Ethics}

The study analyses metadata from published reports. It involves no human
participants, human material or identifiable personal data, and no ethical
approval or informed consent was required.

\section*{Use of generative artificial intelligence}

No generative model was used to screen records, classify studies, extract
data, perform analysis or draft interpretive claims. The characterisation
pipeline uses deterministic pattern matching and metadata lookup only. The
authors take responsibility for all content.

\clearpage
\section*{Tables}
\begin{table}[H]
\centering
\caption{\textbf{Principal estimates.} All quantities are computed on the
11,628 records first available between 2023-Q1 and 2026-Q2. Confidence
intervals are given where the quantity is an estimate rather than a count.
Evaluation lag is defined at Eq.~(\ref{eq:lag}) and the drift offset at
Eq.~(\ref{eq:offset}).}
\label{tab:principal}
\small
\begin{tabularx}{\textwidth}{@{}Xll@{}}
\toprule
\textbf{Quantity} & \textbf{Estimate} & \textbf{95\% CI} \\
\midrule
\multicolumn{3}{@{}l}{\textit{Scale and composition}}\\
Records analysed                                    & 11,628            & --- \\
Growth, incidence rate ratio per quarter            & 1.272             & 1.257--1.289 \\
Records per quarter, 2023-Q1 to 2026-Q2             & 52 to 2,346 (45$\times$) & --- \\
Trial-grade designs                                 & 296 (2.5\%)       & --- \\
\addlinespace
\multicolumn{3}{@{}l}{\textit{Evaluation lag (quarters)}}\\
Mean lag, 2023-Q1 to 2026-Q2                        & 1.33 to 6.08      & --- \\
Observed slope, per quarter                         & 0.329             & --- \\
Counterfactual slope, 2023 composition frozen       & 0.753             & --- \\
Drift offset by migration to newer models           & 56.2\%            & 49.9--65.2 \\
Slope, actively developed families only             & 0.237             & 0.222--0.252 \\
Slope, discontinued families only                   & 1.062             & 1.013--1.110 \\
\addlinespace
\multicolumn{3}{@{}l}{\textit{Study design (difference vs other empirical designs, quarters)}}\\
Randomised trials, median regression                & $+$4.62           & 3.62--5.63 \\
Randomised trials, actively developed families only & $+$0.02           & $-$0.50--0.55 \\
Preprints, actively developed families only         & $-$0.09           & $-$0.31--0.13 \\
Randomised trials evaluating a discontinued family  & 62.1\%            & --- \\
Preprints evaluating a discontinued family          & 15.6\%            & --- \\
\addlinespace
\multicolumn{3}{@{}l}{\textit{Reporting and concentration}}\\
Version specified, randomised trials                & 50.0\%            & --- \\
Version specified, comparative evaluations          & 83.4\%            & --- \\
Herfindahl--Hirschman index, 2023-Q1 to 2026-Q2     & 5,424 to 1,281    & --- \\
Distinct model families named per quarter           & 5 to 37           & --- \\
Purpose-built biomedical models, maximum share      & 2.5\%             & --- \\
\bottomrule
\end{tabularx}
\end{table}

\clearpage
\section*{Figures}

\clearpage
\appendix
\setcounter{page}{1}
\renewcommand{\thepage}{S\arabic{page}}
\begingroup
\titleformat{\section}{\large\bfseries}{}{0pt}{}
\titleformat{\subsection}{\normalsize\bfseries}{}{0pt}{}
\setlength{\parskip}{5pt}

\begin{center}
{\Large\bfseries Supplementary Information}\\[6pt]
{\normalsize The widening evaluation gap in medical large language model
research 2023 to 2026}
\end{center}

\vspace{4pt}\hrule\vspace{10pt}

\textbf{Contents}
\begin{itemize}\setlength{\itemsep}{1pt}
  \item Supplementary Table S1 \textbar{} PRISMA 2020 checklist
  \item Supplementary Table S2 \textbar{} Complete search strings
  \item Supplementary Table S3 \textbar{} Model release table
  \item Supplementary Table S4 \textbar{} Controlled vocabulary
  \item Supplementary Fig. S1 \textbar{} Median evaluation lag by model family
  \item Supplementary Fig. S2 \textbar{} Venue class and journal concentration
\end{itemize}

All derived data and the pipeline that produces them are deposited at
\url{https://doi.org/10.5281/zenodo.21671630}. Tables S2, S3 and S4 are
generated directly from the analysis code, so they cannot diverge from the
analysis as executed.

\clearpage
\section*{Supplementary Table S1 \textbar{} PRISMA 2020 checklist}

This study is a meta-research evidence map. The unit of analysis is the published report, not the patient, and no treatment effect is synthesised. Several PRISMA 2020 items therefore do not apply; each is marked \textbf{N/A} with the reason rather than left blank, as PRISMA advises for reviews that depart from the effect-synthesis model.

\begin{xltabular}{\textwidth}{@{}l>{\raggedright\arraybackslash}p{3.4cm}X@{}}
\toprule
\textbf{\#} & \textbf{Item} & \textbf{Location or reason} \\
\midrule
\endfirsthead
\toprule
\textbf{\#} & \textbf{Item} & \textbf{Location or reason} \\
\midrule
\endhead
\bottomrule
\endfoot
1 & Title identifies the report as a systematic review & \textbf{Partial.} The title identifies the study as an evidence map of a defined literature over a stated window. npj Digital Medicine restricts titles to 15 words free of punctuation, which does not accommodate a design label in addition to the finding. The Abstract and the first line of Methods both state the design. \\
\addlinespace[2pt]
2 & Abstract & Abstract. Structured per PRISMA 2020 for Abstracts as far as the journal's 150-word unstructured limit permits: source, window, corpus size, principal estimates with intervals, and conclusion. \\
\addlinespace[2pt]
3 & Rationale & Introduction, paragraphs 1–3. \\
\addlinespace[2pt]
4 & Objectives & Introduction, final paragraph. Five stated aims. \\
\addlinespace[2pt]
5 & Eligibility criteria & Methods, *Search* and *Analysis window*. Records indexed in PubMed/MEDLINE, first-available date 2023-Q1 to 2026-Q2, retrieved by any of fourteen domain blocks combined with the technology block. No design, language or outcome restriction: reviews and commentaries are retained and characterised, because study design is an analysis variable rather than an exclusion. \\
\addlinespace[2pt]
6 & Information sources & Methods, *Search*. PubMed/MEDLINE via NCBI E-utilities. Date last searched and per-block yields in Supplementary Table S2. \\
\addlinespace[2pt]
7 & Search strategy & Supplementary Table S2, generated directly from the vocabulary module the harvester imports, so the reported and executed searches cannot diverge. \\
\addlinespace[2pt]
8 & Selection process & Methods, *Search* and *Analysis window*; Fig. 1. Eligibility is determined from indexed metadata by deterministic rule, applied by one reviewer with no independent duplicate screening, since the operation is a fixed query rather than a judgement. No full-text retrieval stage applies. \\
\addlinespace[2pt]
9 & Data collection process & Methods, *Record characterisation*. Automated extraction from the PubMed record by fixed pattern and lookup; no manual extraction and therefore no duplicate-extraction step. Rules in Supplementary Table S4. \\
\addlinespace[2pt]
10a & Data items — outcomes & Methods, *Statistical analysis*. Primary: evaluation lag, Eq. (2). Secondary: trial-grade share, version specification, model concentration, attention–evidence gap. \\
\addlinespace[2pt]
10b & Data items — other variables & Methods, *Record characterisation*. Design class, model families named, themes, country and region, venue class, publication quarter. \\
\addlinespace[2pt]
11 & Study risk of bias assessment & \textbf{N/A.} No risk-of-bias appraisal of included records was performed. The study characterises the distribution and currency of the literature; it does not estimate an effect from it, so individual-record validity does not enter any pooled estimate. Stated as a limitation in Discussion. \\
\addlinespace[2pt]
12 & Effect measures & \textbf{N/A.} No effect measure is synthesised across studies. Estimands are properties of the literature and are defined in Eqs. (1)–(9). \\
\addlinespace[2pt]
13a & Synthesis — eligibility for synthesis & Methods, *Statistical analysis*. Records enter each analysis where the relevant variable is defined; coverage of evaluation lag is reported by design class because it is defined only for records naming a model. \\
\addlinespace[2pt]
13b & Synthesis — data preparation & Methods, *Record characterisation* and *Analysis window*. Deduplication on DOI with fallback to PMID; window restriction; handling of unparseable dates. \\
\addlinespace[2pt]
13c & Synthesis — tabulation and visual display & Figs. 1–9; Supplementary Figs. S1–S2. \\
\addlinespace[2pt]
13d & Synthesis — synthesis of results & \textbf{N/A} in the meta-analytic sense. Trends across records are modelled by negative binomial regression, Eq. (1), and median and least-squares regression, Eq. (6), rather than pooled. \\
\addlinespace[2pt]
13e & Synthesis — heterogeneity & \textbf{N/A.} No pooled estimate, so no between-study heterogeneity statistic. Variation across design, domain, region and venue is reported directly. \\
\addlinespace[2pt]
13f & Synthesis — sensitivity analyses & Methods, *Benchmarking the lag trend* and *Design differences*. Pessimistic release assignment; exclusion of encoder-era families; restriction to actively developed families; baseline period varied; Holm correction; bootstrap interval on the drift offset. Results reported in `data/tier1/lag\_robustness.csv`. \\
\addlinespace[2pt]
14 & Reporting bias assessment & \textbf{N/A.} Publication bias concerns the selective reporting of effects. This study counts what was published; unpublished work is outside the estimand by construction. Indexing bias in PubMed is stated as a limitation in Discussion. \\
\addlinespace[2pt]
15 & Certainty assessment & \textbf{N/A.} GRADE applies to certainty in an effect estimate. No effect is estimated. Precision of each descriptive estimate is reported by confidence interval. \\
\addlinespace[2pt]
16a & Study selection results & Results, *Composition of the evidence base*; Fig. 1. 19,406 identified, 7,555 duplicates removed, 11,851 screened, 223 excluded, 11,628 included. \\
\addlinespace[2pt]
16b & Excluded studies & Fig. 1 and `data/tier1/window\_excluded.csv`. Exclusions are by date only: 87 dated outside the window and 136 with no assignable publication date. No record was excluded on judgement. \\
\addlinespace[2pt]
17 & Study characteristics & Results, *Composition of the evidence base* and *Distribution across regions, venues and journals*; Figs. 3, 9. Per-record characteristics deposited. \\
\addlinespace[2pt]
18 & Risk of bias in studies & \textbf{N/A.} See item 11. \\
\addlinespace[2pt]
19 & Results of individual studies & \textbf{N/A.} No individual-study effect estimates are extracted. Per-record derived variables are deposited in `lag\_per\_study.csv`. \\
\addlinespace[2pt]
20a–d & Results of syntheses & Results, throughout. Estimates with 95\% confidence intervals; heterogeneity and certainty items N/A as above. \\
\addlinespace[2pt]
21 & Reporting biases & \textbf{N/A.} See item 14. \\
\addlinespace[2pt]
22 & Certainty of evidence & \textbf{N/A.} See item 15. \\
\addlinespace[2pt]
23a & Interpretation & Discussion, paragraphs 1–3. \\
\addlinespace[2pt]
23b & Limitations of the evidence & Discussion, limitations paragraph. Indexing coverage, metadata-only characterisation, differential coverage of the lag measure. \\
\addlinespace[2pt]
23c & Limitations of review processes & Discussion, limitations paragraph. Single-reviewer deterministic screening; regular-expression model detection; the hand-assigned release table; the n = 22 subgroup underpinning the mechanism claim. \\
\addlinespace[2pt]
23d & Implications & Discussion, final paragraphs. \\
\addlinespace[2pt]
24a & Registration & Methods, *Design and reporting*. Registered on OSF Registries; registration DOI in the Data availability statement. PROSPERO was not used because it does not accept meta-research. \\
\addlinespace[2pt]
24b & Protocol access & Deposited with the data and code. \\
\addlinespace[2pt]
24c & Amendments & Registration was completed after analysis; the record states this explicitly and lists all four deviations from the original analysis plan. \\
\addlinespace[2pt]
25 & Support & Acknowledgements. No specific grant. \\
\addlinespace[2pt]
26 & Competing interests & Competing interests. None declared. \\
\addlinespace[2pt]
27 & Availability of data, code and materials & Data availability and Code availability. Complete pipeline and all derived tables deposited; PubMed records not redistributed, PMIDs and queries supplied. \\
\addlinespace[2pt]
\end{xltabular}

\clearpage
\section*{Supplementary Table S2 \textbar{} Complete search strings}

Emitted directly from \texttt{code/vocab.py}, the module the harvester imports, so this table and the executed search cannot diverge. Each domain block is combined with the single technology block and the date filter \texttt{(\textquotedbl2023/01/01\textquotedbl[dp] : \textquotedbl2026/06/30\textquotedbl[dp])}, then executed separately; results are pooled and deduplicated on DOI with fallback to PMID.

\subsection*{Technology block, constant across all searches}

\noindent \footnotesize \texttt{\seqsplit{("large language model"[tiab] OR "large language models"[tiab] OR "LLM"[tiab] OR "LLMs"[tiab] OR "generative artificial intelligence"[tiab] OR "generative AI"[tiab] OR "foundation model"[tiab] OR "foundation models"[tiab] OR "vision-language model"[tiab] OR "vision language model"[tiab] OR "multimodal large language"[tiab] OR "multimodal model"[tiab] OR "large reasoning model"[tiab] OR "reasoning model"[tiab] OR "instruction-tuned"[tiab] OR "instruction tuned"[tiab] OR "transformer model"[tiab] OR "generative pre-trained"[tiab] OR ChatGPT[tiab] OR "GPT-3"[tiab] OR "GPT-3.5"[tiab] OR "GPT-4"[tiab] OR "GPT-4o"[tiab] OR "GPT-5"[tiab] OR Gemini[tiab] OR "Med-PaLM"[tiab] OR "MedGemma"[tiab] OR Claude[tiab] OR LLaMA[tiab] OR DeepSeek[tiab] OR Qwen[tiab] OR Mistral[tiab] OR Meditron[tiab] OR "OpenBioLLM"[tiab] OR "HuatuoGPT"[tiab])}}

\subsection*{Domain blocks}

\begin{xltabular}{\textwidth}{@{}>{\raggedright\arraybackslash}p{3.6cm}X@{}}
\toprule
\textbf{Domain} & \textbf{Query} \\
\midrule
\endfirsthead
\toprule
\textbf{Domain} & \textbf{Query} \\
\midrule
\endhead
\bottomrule
\endfoot
Medical question answering & \footnotesize \texttt{\seqsplit{("question answering"[tiab] OR USMLE[tiab] OR MedQA[tiab] OR MedMCQA[tiab] OR PubMedQA[tiab] OR MultiMedQA[tiab] OR "MMLU"[tiab] OR "medical licensing"[tiab] OR "board examination"[tiab] OR "board certification"[tiab] OR "multiple choice question"[tiab] OR "knowledge benchmark"[tiab])}} \\
\addlinespace[3pt]
Medical dialogue summarisation & \footnotesize \texttt{\seqsplit{("dialogue summarization"[tiab] OR "dialogue summarisation"[tiab] OR "conversation summarization"[tiab] OR "ambient clinical"[tiab] OR "ambient artificial intelligence"[tiab] OR "ambient documentation"[tiab] OR "AI scribe"[tiab] OR "digital scribe"[tiab] OR "ambient scribe"[tiab] OR "visit note"[tiab] OR "SOAP note"[tiab] OR "encounter note"[tiab])}} \\
\addlinespace[3pt]
EHRs, clinical letters and note generation & \footnotesize \texttt{\seqsplit{("electronic health record"[tiab] OR "electronic health records"[tiab] OR "electronic medical record"[tiab] OR "discharge summary"[tiab] OR "discharge summaries"[tiab] OR "clinical letter"[tiab] OR "clinical documentation"[tiab] OR "note generation"[tiab] OR "documentation burden"[tiab] OR "clinical note"[tiab] OR FHIR[tiab] OR "interoperability"[tiab])}} \\
\addlinespace[3pt]
Scientific research support & \footnotesize \texttt{\seqsplit{("evidence synthesis"[tiab] OR "systematic review"[tiab] OR "literature screening"[tiab] OR "title and abstract screening"[tiab] OR "hypothesis generation"[tiab] OR "scientific discovery"[tiab] OR "laboratory automation"[tiab] OR "data extraction"[tiab] OR "peer review"[tiab] OR "research assistant"[tiab] OR "AI scientist"[tiab] OR "co-scientist"[tiab])}} \\
\addlinespace[3pt]
Medical education and language translation & \footnotesize \texttt{\seqsplit{("medical education"[tiab] OR "medical student"[tiab] OR "medical students"[tiab] OR "nursing education"[tiab] OR "clinical training"[tiab] OR "resident education"[tiab] OR "objective structured clinical"[tiab] OR OSCE[tiab] OR "machine translation"[tiab] OR "medical translation"[tiab] OR "plain language"[tiab] OR "health literacy"[tiab] OR "patient education material"[tiab])}} \\
\addlinespace[3pt]
Medical imaging recognition and reporting & \footnotesize \texttt{\seqsplit{("radiology report"[tiab] OR "radiology reports"[tiab] OR "report generation"[tiab] OR "medical imaging"[tiab] OR "chest radiograph"[tiab] OR "chest x-ray"[tiab] OR "computed tomography"[tiab] OR "magnetic resonance imaging"[tiab] OR "image interpretation"[tiab] OR "medical image"[tiab] OR histopathology[tiab] OR "digital pathology"[tiab] OR dermoscopy[tiab] OR "fundus"[tiab] OR "echocardiograph"[tiab])}} \\
\addlinespace[3pt]
Clinical health and diagnostic reasoning & \footnotesize \texttt{\seqsplit{("diagnostic reasoning"[tiab] OR "clinical reasoning"[tiab] OR "differential diagnosis"[tiab] OR "history taking"[tiab] OR "chain-of-thought"[tiab] OR "chain of thought"[tiab] OR "clinical vignette"[tiab] OR "case vignette"[tiab] OR "diagnostic accuracy"[tiab] OR "cognitive bias"[tiab] OR "anchoring bias"[tiab] OR "test-time compute"[tiab])}} \\
\addlinespace[3pt]
Medical product safety monitoring & \footnotesize \texttt{\seqsplit{(pharmacovigilance[tiab] OR "adverse drug event"[tiab] OR "adverse drug reaction"[tiab] OR "adverse event report"[tiab] OR "signal detection"[tiab] OR "drug safety"[tiab] OR "post-marketing surveillance"[tiab] OR "medication error"[tiab] OR "drug-drug interaction"[tiab] OR MedDRA[tiab] OR "individual case safety report"[tiab])}} \\
\addlinespace[3pt]
Disease diagnosis and prognosis & \footnotesize \texttt{\seqsplit{("disease diagnosis"[tiab] OR "disease classification"[tiab] OR "risk stratification"[tiab] OR "risk prediction"[tiab] OR "prognostic model"[tiab] OR phenotyping[tiab] OR "rare disease"[tiab] OR "early detection"[tiab] OR "screening tool"[tiab] OR "clinical deterioration"[tiab])}} \\
\addlinespace[3pt]
Clinical decision support and administrative tasks & \footnotesize \texttt{\seqsplit{("clinical decision support"[tiab] OR "decision support system"[tiab] OR "medical coding"[tiab] OR "ICD coding"[tiab] OR "billing code"[tiab] OR "prior authorization"[tiab] OR "prior authorisation"[tiab] OR "administrative burden"[tiab] OR "guideline adherence"[tiab] OR "guideline concordance"[tiab] OR "treatment recommendation"[tiab] OR "tumor board"[tiab] OR "care pathway"[tiab])}} \\
\addlinespace[3pt]
Agentic and multi-agent clinical systems & \footnotesize \texttt{\seqsplit{("agentic"[tiab] OR "AI agent"[tiab] OR "AI agents"[tiab] OR "multi-agent"[tiab] OR "multiagent"[tiab] OR "autonomous agent"[tiab] OR "tool use"[tiab] OR "tool calling"[tiab] OR "function calling"[tiab] OR "agent workflow"[tiab] OR "agentic workflow"[tiab] OR "model context protocol"[tiab] OR "orchestration"[tiab])}} \\
\addlinespace[3pt]
Patient-facing conversational health and triage & \footnotesize \texttt{\seqsplit{("symptom checker"[tiab] OR "patient portal"[tiab] OR "patient message"[tiab] OR "patient inquiry"[tiab] OR "health chatbot"[tiab] OR "conversational agent"[tiab] OR "virtual assistant"[tiab] OR "self-triage"[tiab] OR "telehealth"[tiab] OR "telemedicine"[tiab] OR "direct-to-consumer"[tiab] OR "patient-facing"[tiab])}} \\
\addlinespace[3pt]
Mental and behavioural health & \footnotesize \texttt{\seqsplit{("mental health"[tiab] OR "psychiatric"[tiab] OR psychotherapy[tiab] OR "depression"[tiab] OR "anxiety"[tiab] OR "suicide"[tiab] OR "crisis intervention"[tiab] OR "behavioral health"[tiab] OR "behavioural health"[tiab] OR "substance use"[tiab] OR "cognitive behavioral therapy"[tiab] OR "counselling"[tiab] OR "counseling"[tiab])}} \\
\addlinespace[3pt]
Governance, regulation and safety evaluation & \footnotesize \texttt{\seqsplit{("regulatory"[tiab] OR "regulation"[tiab] OR "FDA"[tiab] OR "CE mark"[tiab] OR "AI Act"[tiab] OR "medical device"[tiab] OR "governance"[tiab] OR "red team"[tiab] OR "red-teaming"[tiab] OR "safety evaluation"[tiab] OR "post-deployment monitoring"[tiab] OR "liability"[tiab] OR "accountability"[tiab] OR "informed consent"[tiab] OR "algorithmic audit"[tiab])}} \\
\addlinespace[3pt]
\end{xltabular}

\clearpage
\section*{Supplementary Table S3 \textbar{} Model release table}

Emitted from \texttt{MODEL\_RELEASES} in \texttt{code/23\_evaluation\_lag.py}. Release quarters were assigned by hand from developer announcements and model cards, and are the component of this study most open to challenge; they are reported in full so that any disputed date can be corrected and the analysis re-run. A family is \textbf{active} if its most recent release fell in 2024-Q1 or later. The classification is a property of the family, not of any study, so conditioning on it does not condition on the outcome.

\noindent\textbf{53 families: 24 active, 29 discontinued.}

\begin{xltabular}{\textwidth}{@{}>{\raggedright\arraybackslash}p{2.6cm}lll X@{}}
\toprule
\textbf{Family} & \textbf{First} & \textbf{Latest} & \textbf{Status} & \textbf{All release quarters} \\
\midrule
\endfirsthead
\toprule
\textbf{Family} & \textbf{First} & \textbf{Latest} & \textbf{Status} & \textbf{All release quarters} \\
\midrule
\endhead
\bottomrule
\endfoot
Claude & 2023-Q1 & 2025-Q4 & active & \footnotesize Claude 1 (2023-Q1), Claude 2 (2023-Q3), Claude 3 (2024-Q1), Claude 3.5 (2024-Q2), Claude 4 (2025-Q2), Claude Opus 4.5 (2025-Q4) \\
GPT-5 & 2025-Q3 & 2025-Q4 & active & \footnotesize GPT-5 (2025-Q3), GPT-5.2 (2025-Q4) \\
Gemini & 2023-Q2 & 2025-Q4 & active & \footnotesize PaLM-2/Bard (2023-Q2), Gemini 1.0 (2023-Q4), Gemini 1.5 (2024-Q1), Gemini 2.0 (2024-Q4), Gemini 2.5 (2025-Q1), Gemini 3 (2025-Q4) \\
Baichuan-M & 2025-Q3 & 2025-Q3 & active & \footnotesize Baichuan-M2 (2025-Q3) \\
MedGemma & 2025-Q3 & 2025-Q3 & active & \footnotesize MedGemma (2025-Q3) \\
gpt-oss & 2025-Q3 & 2025-Q3 & active & \footnotesize gpt-oss (2025-Q3) \\
GPT-4 & 2023-Q1 & 2025-Q2 & active & \footnotesize GPT-4 (2023-Q1), GPT-4-turbo (2023-Q4), GPT-4o (2024-Q2), GPT-4.1 (2025-Q2) \\
Llama & 2023-Q1 & 2025-Q2 & active & \footnotesize LLaMA (2023-Q1), Llama 2 (2023-Q3), Llama 3 (2024-Q2), Llama 3.1 (2024-Q3), Llama 4 (2025-Q2) \\
Qwen & 2023-Q3 & 2025-Q2 & active & \footnotesize Qwen (2023-Q3), Qwen2 (2024-Q2), Qwen2.5 (2024-Q3), Qwen3 (2025-Q2) \\
o-series & 2024-Q3 & 2025-Q2 & active & \footnotesize o1 (2024-Q3), o3-mini (2025-Q1), o3 (2025-Q2), o4-mini (2025-Q2) \\
DeepSeek & 2024-Q2 & 2025-Q1 & active & \footnotesize DeepSeek V2 (2024-Q2), DeepSeek V3 (2024-Q4), DeepSeek R1 (2025-Q1) \\
Gemma & 2024-Q1 & 2025-Q1 & active & \footnotesize Gemma (2024-Q1), Gemma 2 (2024-Q2), Gemma 3 (2025-Q1) \\
Grok & 2023-Q4 & 2025-Q1 & active & \footnotesize Grok-1 (2023-Q4), Grok-2 (2024-Q3), Grok-3 (2025-Q1) \\
HuatuoGPT & 2023-Q2 & 2024-Q4 & active & \footnotesize HuatuoGPT (2023-Q2), HuatuoGPT-o1 (2024-Q4) \\
Phi & 2023-Q4 & 2024-Q4 & active & \footnotesize Phi-2 (2023-Q4), Phi-3 (2024-Q2), Phi-4 (2024-Q4) \\
Med-Gemini & 2024-Q2 & 2024-Q2 & active & \footnotesize Med-Gemini (2024-Q2) \\
OpenBioLLM & 2024-Q2 & 2024-Q2 & active & \footnotesize OpenBioLLM (2024-Q2) \\
Apollo-Med & 2024-Q1 & 2024-Q1 & active & \footnotesize Apollo (2024-Q1) \\
BioMistral & 2024-Q1 & 2024-Q1 & active & \footnotesize BioMistral (2024-Q1) \\
CheXagent & 2024-Q1 & 2024-Q1 & active & \footnotesize CheXagent (2024-Q1) \\
Command-R & 2024-Q1 & 2024-Q1 & active & \footnotesize Command R (2024-Q1) \\
GLM & 2023-Q1 & 2024-Q1 & active & \footnotesize ChatGLM (2023-Q1), GLM-4 (2024-Q1) \\
MiniMax & 2024-Q1 & 2024-Q1 & active & \footnotesize MiniMax (2024-Q1) \\
Mistral & 2023-Q3 & 2024-Q1 & active & \footnotesize Mistral 7B (2023-Q3), Mixtral (2023-Q4), Mistral Large (2024-Q1) \\
Kimi & 2023-Q4 & 2023-Q4 & discontinued & \footnotesize Kimi (2023-Q4) \\
MedLM & 2023-Q4 & 2023-Q4 & discontinued & \footnotesize MedLM (2023-Q4) \\
MedSAM & 2023-Q4 & 2023-Q4 & discontinued & \footnotesize MedSAM (2023-Q4) \\
Meditron & 2023-Q4 & 2023-Q4 & discontinued & \footnotesize Meditron (2023-Q4) \\
RadFM & 2023-Q4 & 2023-Q4 & discontinued & \footnotesize RadFM (2023-Q4) \\
Yi & 2023-Q4 & 2023-Q4 & discontinued & \footnotesize Yi (2023-Q4) \\
ClinicalCamel & 2023-Q3 & 2023-Q3 & discontinued & \footnotesize Clinical Camel (2023-Q3) \\
Med-Flamingo & 2023-Q3 & 2023-Q3 & discontinued & \footnotesize Med-Flamingo (2023-Q3) \\
Zhongjing & 2023-Q3 & 2023-Q3 & discontinued & \footnotesize Zhongjing (2023-Q3) \\
Baichuan & 2023-Q2 & 2023-Q2 & discontinued & \footnotesize Baichuan (2023-Q2) \\
ChatDoctor & 2023-Q2 & 2023-Q2 & discontinued & \footnotesize ChatDoctor (2023-Q2) \\
DoctorGLM & 2023-Q2 & 2023-Q2 & discontinued & \footnotesize DoctorGLM (2023-Q2) \\
Falcon & 2023-Q2 & 2023-Q2 & discontinued & \footnotesize Falcon (2023-Q2) \\
InternLM & 2023-Q2 & 2023-Q2 & discontinued & \footnotesize InternLM (2023-Q2) \\
LLaVA-Med & 2023-Q2 & 2023-Q2 & discontinued & \footnotesize LLaVA-Med (2023-Q2) \\
Med-PaLM & 2022-Q4 & 2023-Q2 & discontinued & \footnotesize Med-PaLM (2022-Q4), Med-PaLM 2 (2023-Q2) \\
PMC-LLaMA & 2023-Q2 & 2023-Q2 & discontinued & \footnotesize PMC-LLaMA (2023-Q2) \\
XrayGPT & 2023-Q2 & 2023-Q2 & discontinued & \footnotesize XrayGPT (2023-Q2) \\
BiomedCLIP & 2023-Q1 & 2023-Q1 & discontinued & \footnotesize BiomedCLIP (2023-Q1) \\
GPT-3.5 & 2022-Q4 & 2023-Q1 & discontinued & \footnotesize ChatGPT (2022-Q4), GPT-3.5-turbo (2023-Q1) \\
MedAlpaca & 2023-Q1 & 2023-Q1 & discontinued & \footnotesize MedAlpaca (2023-Q1) \\
BioGPT & 2022-Q3 & 2022-Q3 & discontinued & \footnotesize BioGPT (2022-Q3) \\
GatorTron & 2022-Q2 & 2022-Q2 & discontinued & \footnotesize GatorTron (2022-Q2) \\
PubMedBERT & 2020-Q3 & 2020-Q3 & discontinued & \footnotesize PubMedBERT (2020-Q3) \\
GPT-3 & 2020-Q2 & 2020-Q2 & discontinued & \footnotesize GPT-3 (2020-Q2) \\
T5 & 2019-Q4 & 2019-Q4 & discontinued & \footnotesize T5 (2019-Q4) \\
BioBERT & 2019-Q3 & 2019-Q3 & discontinued & \footnotesize BioBERT (2019-Q3) \\
ClinicalBERT & 2019-Q2 & 2019-Q2 & discontinued & \footnotesize ClinicalBERT (2019-Q2) \\
BERT & 2018-Q4 & 2018-Q4 & discontinued & \footnotesize BERT (2018-Q4) \\
\end{xltabular}

\clearpage
\section*{Supplementary Table S4 \textbar{} Controlled vocabulary}

Emitted from \texttt{code/vocab.py}. Every rule is a fixed pattern or lookup applied identically to all records; no generative model is used at any stage of screening, classification or extraction.

\subsection*{S4.1 Model-family detection (53 patterns)}

Applied to the concatenated title and abstract, case-insensitive.

\begin{xltabular}{\textwidth}{@{}>{\raggedright\arraybackslash}p{2.4cm}>{\raggedright\arraybackslash}p{2.6cm}X@{}}
\toprule
\textbf{Family} & \textbf{Class} & \textbf{Regular expression} \\
\midrule
\endfirsthead
\toprule
\textbf{Family} & \textbf{Class} & \textbf{Regular expression} \\
\midrule
\endhead
\bottomrule
\endfoot
Apollo-Med & biomedical specialist & \footnotesize \texttt{\seqsplit{\textbackslash{}bapollo\textbackslash{}b(?=.\{0,60\}\textbackslash{}bmedical\textbackslash{}b)}} \\
BERT & extractive baseline & \footnotesize \texttt{\seqsplit{(?<!pubmed)(?<!bio)(?<!clinical)(?<!sci)\textbackslash{}bbert\textbackslash{}b|\textbackslash{}broberta\textbackslash{}b|\textbackslash{}bdeberta\textbackslash{}b}} \\
Baichuan & open-weight generalist & \footnotesize \texttt{\seqsplit{\textbackslash{}bbaichuan\textbackslash{}b(?![-\textbackslash{}s]?m)}} \\
Baichuan-M & biomedical specialist & \footnotesize \texttt{\seqsplit{\textbackslash{}bbaichuan[-\textbackslash{}s]?m\textbackslash{}d?\textbackslash{}b}} \\
BioBERT & extractive baseline & \footnotesize \texttt{\seqsplit{\textbackslash{}bbiobert\textbackslash{}b}} \\
BioGPT & biomedical specialist & \footnotesize \texttt{\seqsplit{\textbackslash{}bbiogpt\textbackslash{}b}} \\
BioMistral & biomedical specialist & \footnotesize \texttt{\seqsplit{\textbackslash{}bbiomistral\textbackslash{}b}} \\
BiomedCLIP & biomedical multimodal & \footnotesize \texttt{\seqsplit{\textbackslash{}bbiomedclip\textbackslash{}b}} \\
ChatDoctor & biomedical specialist & \footnotesize \texttt{\seqsplit{\textbackslash{}bchatdoctor\textbackslash{}b}} \\
CheXagent & biomedical multimodal & \footnotesize \texttt{\seqsplit{\textbackslash{}bchexagent\textbackslash{}b}} \\
Claude & closed generalist & \footnotesize \texttt{\seqsplit{\textbackslash{}bclaude\textbackslash{}b|\textbackslash{}bsonnet\textbackslash{}b|\textbackslash{}bopus\textbackslash{}b(?!\textbackslash{}s*\textbackslash{}d*\textbackslash{}s*magnum)}} \\
ClinicalBERT & extractive baseline & \footnotesize \texttt{\seqsplit{\textbackslash{}bclinicalbert\textbackslash{}b|\textbackslash{}bbio\_?clinicalbert\textbackslash{}b}} \\
ClinicalCamel & biomedical specialist & \footnotesize \texttt{\seqsplit{\textbackslash{}bclinical ?camel\textbackslash{}b}} \\
Command-R & closed generalist & \footnotesize \texttt{\seqsplit{\textbackslash{}bcommand[-\textbackslash{}s]?r\textbackslash{}+?\textbackslash{}b}} \\
DeepSeek & open-weight generalist & \footnotesize \texttt{\seqsplit{\textbackslash{}bdeepseek\textbackslash{}b}} \\
DoctorGLM & biomedical specialist & \footnotesize \texttt{\seqsplit{\textbackslash{}bdoctorglm\textbackslash{}b}} \\
Falcon & open-weight generalist & \footnotesize \texttt{\seqsplit{\textbackslash{}bfalcon[-\textbackslash{}s]?\textbackslash{}d*b?\textbackslash{}b}} \\
GLM & open-weight generalist & \footnotesize \texttt{\seqsplit{\textbackslash{}bchatglm\textbackslash{}b|\textbackslash{}bglm[-\textbackslash{}s]?\textbackslash{}d\textbackslash{}b}} \\
GPT-3 & closed generalist & \footnotesize \texttt{\seqsplit{\textbackslash{}bgpt[-\textbackslash{}s]?3\textbackslash{}b(?!\textbackslash{}.5)|\textbackslash{}binstructgpt\textbackslash{}b}} \\
GPT-3.5 & closed generalist & \footnotesize \texttt{\seqsplit{\textbackslash{}bgpt[-\textbackslash{}s]?3\textbackslash{}.5\textbackslash{}b|\textbackslash{}bchatgpt\textbackslash{}b}} \\
GPT-4 & closed generalist & \footnotesize \texttt{\seqsplit{\textbackslash{}bgpt[-\textbackslash{}s]?4(?:o|v|\textbackslash{}.\textbackslash{}d|[-\textbackslash{}s]turbo)?\textbackslash{}b}} \\
GPT-5 & closed generalist & \footnotesize \texttt{\seqsplit{\textbackslash{}bgpt[-\textbackslash{}s]?5(?:\textbackslash{}.\textbackslash{}d)?\textbackslash{}b}} \\
GatorTron & biomedical specialist & \footnotesize \texttt{\seqsplit{\textbackslash{}bgatortron\textbackslash{}b}} \\
Gemini & closed generalist & \footnotesize \texttt{\seqsplit{\textbackslash{}bgemini\textbackslash{}b|\textbackslash{}bbard\textbackslash{}b|\textbackslash{}bpalm[-\textbackslash{}s]?2\textbackslash{}b}} \\
Gemma & open-weight generalist & \footnotesize \texttt{\seqsplit{(?<!med)\textbackslash{}bgemma[-\textbackslash{}s]?\textbackslash{}d?\textbackslash{}b}} \\
Grok & closed generalist & \footnotesize \texttt{\seqsplit{\textbackslash{}bgrok\textbackslash{}b}} \\
HuatuoGPT & biomedical specialist & \footnotesize \texttt{\seqsplit{\textbackslash{}bhuatuo ?gpt\textbackslash{}b|\textbackslash{}bhuatuo\textbackslash{}b|\textbackslash{}bbentsao\textbackslash{}b}} \\
InternLM & open-weight generalist & \footnotesize \texttt{\seqsplit{\textbackslash{}binternlm\textbackslash{}b}} \\
Kimi & open-weight generalist & \footnotesize \texttt{\seqsplit{\textbackslash{}bkimi\textbackslash{}b}} \\
LLaVA-Med & biomedical multimodal & \footnotesize \texttt{\seqsplit{\textbackslash{}bllava[-\textbackslash{}s]?med\textbackslash{}b}} \\
Llama & open-weight generalist & \footnotesize \texttt{\seqsplit{\textbackslash{}bllama[-\textbackslash{}s]?\textbackslash{}d?\textbackslash{}b|\textbackslash{}balpaca\textbackslash{}b|\textbackslash{}bvicuna\textbackslash{}b}} \\
Med-Flamingo & biomedical multimodal & \footnotesize \texttt{\seqsplit{\textbackslash{}bmed[-\textbackslash{}s]?flamingo\textbackslash{}b}} \\
Med-Gemini & biomedical specialist & \footnotesize \texttt{\seqsplit{\textbackslash{}bmed[-\textbackslash{}s]?gemini\textbackslash{}b}} \\
Med-PaLM & biomedical specialist & \footnotesize \texttt{\seqsplit{\textbackslash{}bmed[-\textbackslash{}s]?palm\textbackslash{}b}} \\
MedAlpaca & biomedical specialist & \footnotesize \texttt{\seqsplit{\textbackslash{}bmedalpaca\textbackslash{}b}} \\
MedGemma & biomedical specialist & \footnotesize \texttt{\seqsplit{\textbackslash{}bmedgemma\textbackslash{}b|\textbackslash{}bmedsiglip\textbackslash{}b}} \\
MedLM & biomedical specialist & \footnotesize \texttt{\seqsplit{\textbackslash{}bmedlm\textbackslash{}b}} \\
MedSAM & biomedical multimodal & \footnotesize \texttt{\seqsplit{\textbackslash{}bmedsam\textbackslash{}b}} \\
Meditron & biomedical specialist & \footnotesize \texttt{\seqsplit{\textbackslash{}bmeditron\textbackslash{}b}} \\
MiniMax & open-weight generalist & \footnotesize \texttt{\seqsplit{\textbackslash{}bminimax\textbackslash{}b}} \\
Mistral & open-weight generalist & \footnotesize \texttt{\seqsplit{\textbackslash{}bmistral\textbackslash{}b|\textbackslash{}bmixtral\textbackslash{}b}} \\
OpenBioLLM & biomedical specialist & \footnotesize \texttt{\seqsplit{\textbackslash{}bopenbio ?llm\textbackslash{}b}} \\
PMC-LLaMA & biomedical specialist & \footnotesize \texttt{\seqsplit{\textbackslash{}bpmc[-\textbackslash{}s]?llama\textbackslash{}b}} \\
Phi & open-weight generalist & \footnotesize \texttt{\seqsplit{\textbackslash{}bphi[-\textbackslash{}s]?[2345]\textbackslash{}b}} \\
PubMedBERT & extractive baseline & \footnotesize \texttt{\seqsplit{\textbackslash{}bpubmedbert\textbackslash{}b|\textbackslash{}bbiomedbert\textbackslash{}b}} \\
Qwen & open-weight generalist & \footnotesize \texttt{\seqsplit{\textbackslash{}bqwen\textbackslash{}b}} \\
RadFM & biomedical multimodal & \footnotesize \texttt{\seqsplit{\textbackslash{}bradfm\textbackslash{}b}} \\
T5 & extractive baseline & \footnotesize \texttt{\seqsplit{\textbackslash{}bt5\textbackslash{}b|\textbackslash{}bflan[-\textbackslash{}s]?t5\textbackslash{}b|\textbackslash{}bscifive\textbackslash{}b}} \\
XrayGPT & biomedical multimodal & \footnotesize \texttt{\seqsplit{\textbackslash{}bxray ?gpt\textbackslash{}b}} \\
Yi & open-weight generalist & \footnotesize \texttt{\seqsplit{\textbackslash{}byi[-\textbackslash{}s]?(?:6b|34b|large)\textbackslash{}b}} \\
Zhongjing & biomedical specialist & \footnotesize \texttt{\seqsplit{\textbackslash{}bzhongjing\textbackslash{}b}} \\
gpt-oss & open-weight generalist & \footnotesize \texttt{\seqsplit{\textbackslash{}bgpt[-\textbackslash{}s]?oss\textbackslash{}b}} \\
o-series & closed generalist & \footnotesize \texttt{\seqsplit{\textbackslash{}bopenai\textbackslash{}s+o[134]\textbackslash{}b|\textbackslash{}bo1[-\textbackslash{}s](?:preview|mini)\textbackslash{}b|\textbackslash{}bo3[-\textbackslash{}s]?mini\textbackslash{}b|\textbackslash{}bo4[-\textbackslash{}s]?mini\textbackslash{}b}} \\
\end{xltabular}

\subsection*{S4.2 Themes (18)}

\begin{xltabular}{\textwidth}{@{}>{\raggedright\arraybackslash}p{4.1cm}X@{}}
\toprule
\textbf{Theme} & \textbf{Matching terms} \\
\midrule
\endfirsthead
\toprule
\textbf{Theme} & \textbf{Matching terms} \\
\midrule
\endhead
\bottomrule
\endfoot
adaptation & \footnotesize \texttt{\seqsplit{fine[-\textbackslash{}s]?tun|instruction[-\textbackslash{}s]?tun|\textbackslash{}blora\textbackslash{}b|\textbackslash{}bpeft\textbackslash{}b|parameter[-\textbackslash{}s]?efficient|domain adaptation|continued pretrain|distillat|quantiz|quantis}} \\
agentic & \footnotesize \texttt{\seqsplit{agentic|\textbackslash{}bai agent|multi[-\textbackslash{}s]?agent|autonomous agent|tool[-\textbackslash{}s]?(?:use|call)|function calling|model context protocol}} \\
benchmark\_validity & \footnotesize \texttt{\seqsplit{benchmark|leaderboard|contaminat|data leakage|construct validity|llm[-\textbackslash{}s]?as[-\textbackslash{}s]?(?:a[-\textbackslash{}s]?)?judge}} \\
calibration & \footnotesize \texttt{\seqsplit{calibrat|uncertainty|confidence estimat|abstention|selective prediction}} \\
deployment & \footnotesize \texttt{\seqsplit{deploy|implementation|real[-\textbackslash{}s]?world|prospective|workflow integration|pilot|scale[-\textbackslash{}s]?up|adoption|usability|workload|burnout}} \\
economics & \footnotesize \texttt{\seqsplit{cost|cost[-\textbackslash{}s]?effectiv|latency|throughput|inference cost|token budget|resource constrain|edge deployment|on[-\textbackslash{}s]?device}} \\
equity & \footnotesize \texttt{\seqsplit{\textbackslash{}bbias\textbackslash{}b|fairness|disparit|equity|underserved|low[-\textbackslash{}s]?resource|multilingual|non[-\textbackslash{}s]?english|health equity}} \\
explainability & \footnotesize \texttt{\seqsplit{explainab|interpretab|transparen|attribution|saliency|rationale}} \\
explicit\_reasoning & \footnotesize \texttt{\seqsplit{chain[-\textbackslash{}s]?of[-\textbackslash{}s]?thought|step[-\textbackslash{}s]?by[-\textbackslash{}s]?step reasoning|test[-\textbackslash{}s]?time compute|extended thinking|reasoning trace|self[-\textbackslash{}s]?consistency|reflection}} \\
hallucination & \footnotesize \texttt{\seqsplit{hallucinat|confabulat|factual(?:ity| error)|fabricat|omission|groundedness}} \\
human\_oversight & \footnotesize \texttt{\seqsplit{human[-\textbackslash{}s]?in[-\textbackslash{}s]?the[-\textbackslash{}s]?loop|clinician review|oversight|automation bias|verification|supervis|physician[-\textbackslash{}s]?in[-\textbackslash{}s]?the[-\textbackslash{}s]?loop}} \\
knowledge\_grounding & \footnotesize \texttt{\seqsplit{knowledge graph|ontolog|\textbackslash{}bumls\textbackslash{}b|\textbackslash{}bsnomed\textbackslash{}b|\textbackslash{}bicd[-\textbackslash{}s]?1[01]\textbackslash{}b|terminolog|graph[-\textbackslash{}s]?rag}} \\
multimodal & \footnotesize \texttt{\seqsplit{multimodal|vision[-\textbackslash{}s]?language|image[-\textbackslash{}s]?text|volumetric|3d\textbackslash{}b|voxel}} \\
privacy & \footnotesize \texttt{\seqsplit{privacy|de[-\textbackslash{}s]?identif|\textbackslash{}bhipaa\textbackslash{}b|\textbackslash{}bgdpr\textbackslash{}b|federated|differential privacy|synthetic data|re[-\textbackslash{}s]?identif}} \\
prompting & \footnotesize \texttt{\seqsplit{prompt engineer|zero[-\textbackslash{}s]?shot|few[-\textbackslash{}s]?shot|in[-\textbackslash{}s]?context learning|prompt design|prompt template}} \\
regulation & \footnotesize \texttt{\seqsplit{regulat|\textbackslash{}bfda\textbackslash{}b|ce[-\textbackslash{}s]?mark|ai act|medical device|governance|liabilit|accountab|certification}} \\
retrieval\_augmentation & \footnotesize \texttt{\seqsplit{retrieval[-\textbackslash{}s]?augmented|\textbackslash{}brag\textbackslash{}b|retrieval augmentation|vector (?:data)?base|semantic search}} \\
safety\_guardrails & \footnotesize \texttt{\seqsplit{guardrail|red[-\textbackslash{}s]?team|adversarial|jailbreak|prompt injection|safety filter|refusal|misuse}} \\
\end{xltabular}

\subsection*{S4.3 Design classification}

PubMed PublicationType tags are mapped to analysis classes in the priority order below; the first match wins, so a record tagged both as a randomised controlled trial and as a journal article is classed as the former.

\begin{longtable}{@{}rll@{}}
\toprule
\textbf{Priority} & \textbf{Publication type} & \textbf{Analysis class} \\
\midrule
\endfirsthead
\toprule
\textbf{Priority} & \textbf{Publication type} & \textbf{Analysis class} \\
\midrule
\endhead
\bottomrule
\endfoot
1 & Randomized Controlled Trial & randomised controlled trial \\
2 & Pragmatic Clinical Trial & randomised controlled trial \\
3 & Equivalence Trial & randomised controlled trial \\
4 & Clinical Trial, Phase IV & clinical trial \\
5 & Clinical Trial, Phase III & clinical trial \\
6 & Clinical Trial, Phase II & clinical trial \\
7 & Clinical Trial, Phase I & clinical trial \\
8 & Controlled Clinical Trial & clinical trial \\
9 & Clinical Trial Protocol & trial protocol \\
10 & Clinical Trial & clinical trial \\
11 & Observational Study & observational study \\
12 & Multicenter Study & observational study \\
13 & Comparative Study & comparative evaluation \\
14 & Validation Study & validation study \\
15 & Evaluation Study & comparative evaluation \\
16 & Systematic Review & review \\
17 & Meta-Analysis & review \\
18 & Scoping Review & review \\
19 & Review & review \\
20 & Editorial & commentary \\
21 & Comment & commentary \\
22 & Letter & commentary \\
23 & News & commentary \\
24 & Preprint & preprint \\
\end{longtable}

\noindent\textbf{Trial-grade classes} (randomised, controlled or prospective): clinical trial, observational study, randomised controlled trial.

\subsection*{S4.4 Clinical domains (14)}

Full search strings are given in Supplementary Table S2.

\begin{longtable}{@{}l>{\raggedright\arraybackslash}p{7.6cm}l@{}}
\toprule
\textbf{Key} & \textbf{Domain} & \textbf{Provenance} \\
\midrule
\endfirsthead
\toprule
\textbf{Key} & \textbf{Domain} & \textbf{Provenance} \\
\midrule
\endhead
\bottomrule
\endfoot
MQA & Medical question answering & Wang and Zhang (2024) \\
MDS & Medical dialogue summarisation & Wang and Zhang (2024) \\
EHR & EHRs, clinical letters and note generation & Wang and Zhang (2024) \\
SCI & Scientific research support & Wang and Zhang (2024) \\
EDU & Medical education and language translation & Wang and Zhang (2024) \\
IMG & Medical imaging recognition and reporting & Wang and Zhang (2024) \\
CDR & Clinical health and diagnostic reasoning & Wang and Zhang (2024) \\
PSM & Medical product safety monitoring & Wang and Zhang (2024) \\
DDX & Disease diagnosis and prognosis & Wang and Zhang (2024) \\
CDS & Clinical decision support and administrative tasks & Wang and Zhang (2024) \\
AGT & Agentic and multi-agent clinical systems & added here \\
PFC & Patient-facing conversational health and triage & added here \\
MBH & Mental and behavioural health & added here \\
GOV & Governance, regulation and safety evaluation & added here \\
\end{longtable}

\clearpage
\section*{Supplementary Fig. S1 \textbar{} Median evaluation lag by model
family}

\begin{figure}[H]\centering
\includegraphics[width=0.72\textwidth]{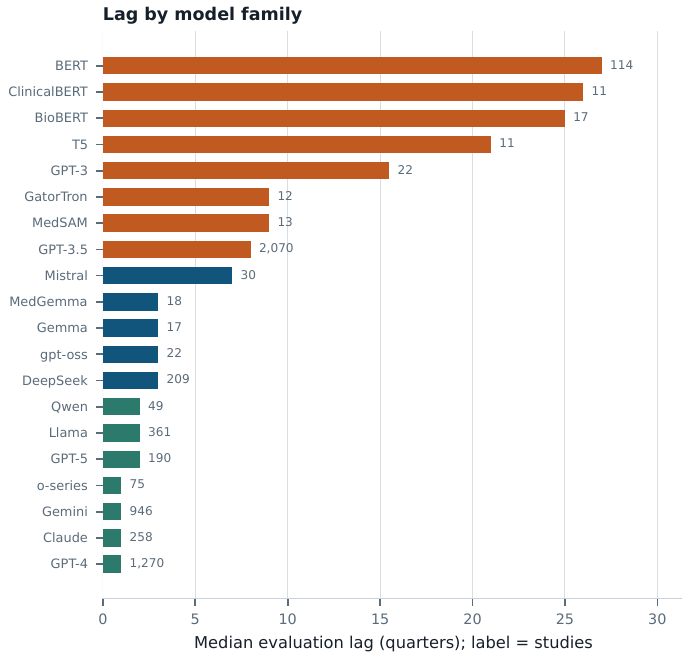}
\caption*{Median evaluation lag for every model family named by at least ten
records, with the number of studies labelled. Families whose development had
ceased before the analysis window carry the longest lags; those still
receiving releases cluster at one to three quarters.}
\end{figure}

\clearpage
\section*{Supplementary Fig. S2 \textbar{} Venue class and journal
concentration}

\begin{figure}[H]\centering
\includegraphics[width=0.86\textwidth]{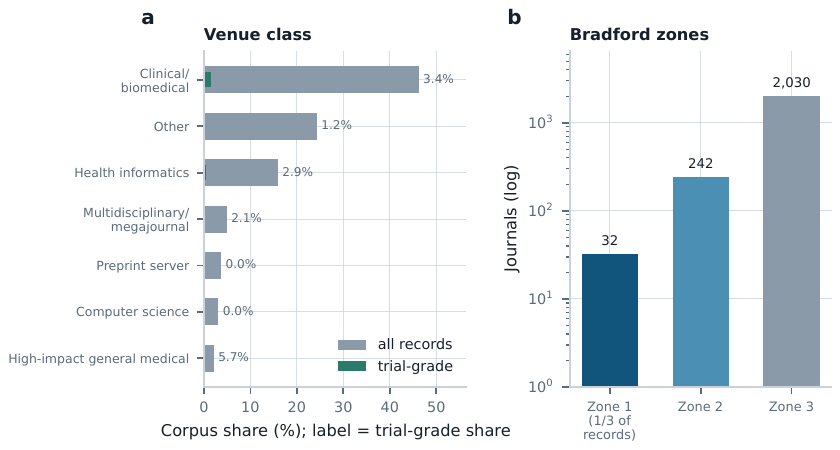}
\caption*{\textbf{a}, Corpus share by venue class, with the trial-grade share
within each class labelled. \textbf{b}, Bradford zones: the number of journals
carrying each successive third of the corpus, on a logarithmic scale.}
\end{figure}

\endgroup

\end{document}